\documentclass[10pt,journal,compsoc]{IEEEtran}
\ifCLASSOPTIONcompsoc
\usepackage[nocompress]{cite}
  
\else
  \usepackage{cite}
\fi

\ifCLASSINFOpdf
  \usepackage{graphicx}
\else
\fi

\usepackage{amsmath,amsfonts}
\usepackage{algorithm}
\usepackage{algorithmic}
\usepackage{color}
\usepackage{multirow}
\usepackage{amssymb}
\usepackage{caption}
\usepackage{orcidlink}
\usepackage{graphicx}
\usepackage{ragged2e}
\usepackage{booktabs}
\usepackage{textcomp}
\usepackage{stfloats}
\usepackage{colortbl}

\usepackage{url}
\usepackage{verbatim}
\usepackage{pifont}
\usepackage{arydshln}
\usepackage{comment}
\usepackage{enumitem}
\usepackage{listings}
\definecolor{lightgray}{gray}{0.9}
\usepackage{mydef}
\usepackage{hyperref}
\hypersetup{colorlinks=true,linkcolor=blue,citecolor=blue,urlcolor=blue}

\definecolor{citecolor}{HTML}{2fe062}

\usepackage{wrapfig}
\usepackage[normalem]{ulem}
\usepackage{makecell}
\usepackage{subcaption}
\usepackage{multirow}
\usepackage{diagbox}
\usepackage[etex=true,export]{adjustbox}
\usepackage{bbding}
\usepackage[edges]{forest}
\definecolor{hidden-draw}{RGB}{106,142,189}
\definecolor{hidden-blue}{RGB}{135,206,250}
\definecolor{hidden-orange}{RGB}{217, 232, 252}
\usepackage{tabularx}

\lstdefinestyle{mystyle}{
    language=Python,
    basicstyle=\ttfamily\small,
    keywordstyle=\color{black},
    commentstyle=\color{black},
    stringstyle=\color{black},
    breakatwhitespace=false,         
    breaklines=true,                 
    captionpos=b,                    
    keepspaces=true,                 
    showspaces=false,                
    showstringspaces=false,
    showtabs=false,                  
    tabsize=2
}

\usepackage{balance}

\usepackage{academicons}
\usepackage{etoolbox}

\usepackage{todonotes}\usepackage{etoolbox}
\begin{document}

\title{Parameter-Efficient Fine-Tuning for Pre-Trained Vision Models: A Survey and Benchmark}

\author{
Yi~Xin, 
Jianjiang Yang, 
Siqi~Luo,
Yuntao~Du,
Qi Qin,
Kangrui Cen,
Yangfan He,
Zhiwei Zhang,
Bin Fu,\\
Xiaokang~Yang, 
Guangtao~Zhai, 
Ming-Hsuan Yang, 
Xiaohong~Liu 
\IEEEcompsocitemizethanks{\IEEEcompsocthanksitem Yi Xin is with the Nanjing University and Shanghai Innovation Institute. 
\IEEEcompsocthanksitem
Siqi Luo, Kangrui Cen, Guangtao Zhai, Xiaokang Yang, Xiaohong Liu are with the Shanghai Jiao Tong University. 
\IEEEcompsocthanksitem Jianjiang Yang is with the University of Bristol.
\IEEEcompsocthanksitem Yuntao Du is with the Shandong University.
\IEEEcompsocthanksitem Qi Qin is with the University of Sydney.
\IEEEcompsocthanksitem Yangfan He is with the University of Minnesota Twin Cities. 
\IEEEcompsocthanksitem Zhiwei Zhang is with The Pennsylvania State University. 
\IEEEcompsocthanksitem Bin Fu is with the Shenzhen Institutes of Advanced Technology, Chinese Academy of Sciences.

\IEEEcompsocthanksitem Ming-Hsuan Yang is with the University of California at Merced.

\IEEEcompsocthanksitem First three authors contributed equally.

\IEEEcompsocthanksitem Corresponding Author: Xiaohong Liu. E-mail: xiaohongliu@sjtu.edu.cn.

\IEEEcompsocthanksitem A preliminary version of the benchmark has been presented in the NeurIPS 2024~\cite{xin2024v}.
}

}

\IEEEtitleabstractindextext{%
\begin{abstract}
\justifying  
Pre-trained vision models (PVMs) have demonstrated remarkable adaptability across a wide range of downstream vision tasks, showcasing exceptional performance. However, as these models scale to billions or even trillions of parameters, conventional full fine-tuning has become increasingly impractical due to its high computational and storage demands. To address these challenges, parameter-efficient fine-tuning (PEFT) has emerged as a promising alternative, aiming to achieve performance comparable to full fine-tuning while making minimal adjustments to the model parameters. This paper presents a comprehensive survey of the latest advancements in the visual PEFT field, systematically reviewing current methodologies and categorizing them into four primary categories: addition-based, partial-based, unified-based, and multi-task tuning. In addition, this paper offers an in-depth analysis of widely used visual datasets and real-world applications where PEFT methods have been successfully applied. Furthermore, this paper introduces the V-PEFT Bench, a unified benchmark designed to standardize the evaluation of PEFT methods across a diverse set of vision tasks, ensuring consistency and fairness in comparison. Finally, the paper outlines potential directions for future research to propel advances in the PEFT field.
A comprehensive collection of resources is available at \url{https://github.com/synbol/Awesome-Parameter-Efficient-Transfer-Learning}, and the benchmark codebase is available at \url{https://github.com/synbol/Parameter-Efficient-Transfer-Learning-Benchmark}.
\end{abstract}

}

\maketitle

\IEEEdisplaynontitleabstractindextext

\IEEEpeerreviewmaketitle

\section{Introduction}
\label{sec:introduction}
\IEEEPARstart{W}{ith} rapid advances in visual datasets~\cite{deng2009imagenet,ridnik2021imagenet,kay2017kinetics}, model architectures~\cite{he2016deep,liu2021swin,Wang2021PyramidVT,kirillov2023segment}, and training algorithms~\cite{touvron2021training,he2022masked,chen2020simple,he2020momentum}, a significant number of vision foundation models have been developed. 
In particular, transformer-based pre-trained vision models (PVMs) have demonstrated remarkable performance across a wide range of computer vision (CV) tasks, such as image recognition~\cite{dosovitskiy2020vit,hou2024conv2former,li2022contextual}, video action recognition~\cite{yang2022recurring,xing2023svformer}, semantic segmentation~\cite{strudel2021segmenter,zhang2022segvit,kirillov2023segment}, and image generation~\cite{chen2023pixart,liu2024lumina,qin2025lumina}, among others.

Due to the impressive representational capabilities of PVMs, it has emerged as a dominant approach to fine-tune PVMs to address specific downstream tasks. 
However, conventional full fine-tuning, while effective, demands substantial computational and memory resources. 
This becomes particularly expensive when dealing with models that contain billions or even trillions of parameters. 
Furthermore, in cases where downstream datasets are small, the model is prone to overfitting, leading to poor generalization performance. 
Additionally, maintaining a separate set of model weights for each task becomes increasingly impractical as the number of tasks grows, particularly for large-scale PVMs.

To address these challenges, parameter-efficient fine-tuning (PEFT) has emerged as a promising solution. 
Originally introduced in the field of natural language processing (NLP)~\cite{houlsby2019parameter}, PEFT mitigates the drawbacks of conventional full fine-tuning by focusing on updating only a small subset of parameters. 
This approach has the potential to match or even exceed the performance of full fine-tuning, while requiring significantly fewer computational and memory resources. 
The effectiveness of PEFT is supported by recent advancements that demonstrate the strong generalizability of large pre-trained models trained on diverse high-quality datasets~\cite{kornblith2019better,yu2022towards}. 
These studies indicate that most of the parameters in PVMs can be shared across a variety of downstream tasks without compromising performance. 
By reducing the number of learnable parameters, PEFT enables more efficient adaptation to new tasks while preserving the valuable pre-existing knowledge embedded in PVMs.

\begin{table*}[thbp]
\centering
\setlength{\tabcolsep}{4pt} 
\renewcommand{\arraystretch}{1} 
\vspace{-0.1cm}
\caption{\textbf{Existing Parameter-Efficient Fine-Tuning Surveys.} The comparison is based on their method taxonomy, domain coverage, and benchmark support.}
\resizebox{0.95\linewidth}{!}{
\begin{tabular}{@{}ccccccccc@{}}
\toprule
\multirow{2}{*}{\textbf{Survey}} & \multicolumn{4}{c}{\textbf{Taxonomy}} &  \multicolumn{3}{c}{\textbf{Domain}} & \multirow{2}{*}{\textbf{Benchmark}}\\
\cmidrule(lr){2-5}\cmidrule(lr){6-8}
 & \makecell[c]{Addition-based} & \makecell[c]{Partial-based} & \makecell[c]{Unified-based} & \makecell[c]{Multi-task} & \makecell[c]{NLP} & \makecell[c]{CV} &\makecell[c]{Diffusion Model} \\
\midrule
\rowcolor{cyan!8}\cite{ding2022delta} & \checkmark & \checkmark & & &\checkmark &\cellcolor{cyan!8}&\cellcolor{cyan!8}&\cellcolor{cyan!8}\\
\cite{xu2023parameter} & \checkmark & \checkmark &\checkmark &\checkmark  &\checkmark & &\\
\cellcolor{cyan!8}\cite{han2024parameter} & \cellcolor{cyan!8}\checkmark & \cellcolor{cyan!8}\checkmark &\cellcolor{cyan!8}\checkmark &\cellcolor{cyan!8}&\cellcolor{cyan!8}\checkmark &\cellcolor{cyan!8}&\cellcolor{cyan!8}\checkmark &\cellcolor{cyan!8}\\
\cite{wang2024parameter} & \checkmark & \checkmark & &\checkmark &\checkmark & &\checkmark\\
\cellcolor{cyan!8}Ours & \cellcolor{cyan!8}\checkmark & \cellcolor{cyan!8}\checkmark &\cellcolor{cyan!8}\checkmark &\cellcolor{cyan!8}\checkmark  &\cellcolor{cyan!8}&\cellcolor{cyan!8}\checkmark &\cellcolor{cyan!8}\checkmark &\cellcolor{cyan!8}\checkmark\\
\bottomrule
\end{tabular}
}
\label{tab:surveys}
\vspace{-0.3cm}
\end{table*}

Given the rapid development of large-scale PVMs and the increasing importance of visual PEFT strategies, there is a growing need for comprehensive surveys that provide detailed and up-to-date insights into the application of PEFT in the vision domain. 
In addition, numerous PEFT methods have been proposed for vision tasks, their direct comparison is not yet standardized. 
In contrast, the NLP domain has developed integrated libraries, which consolidate various PEFT methods and large language models (LLMs) to facilitate their use in downstream tasks. 
Therefore, it is desired to develop a library for the vision domain, which could boost the development of PEFT.

In response to the urgent challenges in the visual PEFT field outlined above, our contributions are as follows: 
\begin{itemize} 
    \item In Section~\ref{sec:preliminary}, we provide a comprehensive definition of the PEFT problem, outlining some of the most widely used vision foundation models and exploring the various pre-training techniques that have contributed significantly to the development of PEFT.

    \item In Section~\ref{sec:taxonomy}, we present a systematic and thorough survey of PEFT methods within the visual domain, with a particular focus on transformer-based approaches. The existing visual PEFT techniques are categorized into addition-based tuning, partial-based tuning, unified-based tuning, and multi-task tuning.

    \item In Section~\ref{task_dataset}, we analyze the prominent datasets and applications within the visual PEFT domain. These datasets and applications are diverse and encompass nearly all major visual tasks.
    
    \item In Section~\ref{sec:benchmark}, we introduce the V-PEFT Bench, a unified and rigorous benchmark for evaluating PEFT methods in visual tasks, ensuring fair and consistent assessments. Within V-PEFT Bench, we propose a Performance-Parameter Trade-off (PPT) metric and conduct evaluations on 2 conventional methods and 25 PEFT methods across 24 image recognition datasets, 3 video action recognition datasets, and 3 dense prediction tasks.

    \item Following an in-depth review and benchmarking of existing PEFT methods, we propose several innovative and forward-looking research directions. These include considerations on the interpretability of PEFT methods, advances in the field of visual generation, and strategies for simplifying hyperparameter selection, as discussed in Section~\ref{sec:direction}.
\end{itemize}

\vspace{1mm}
\noindent \textbf{Differences from Existing Surveys.}
We compare our survey with several existing PEFT surveys in terms of method taxonomy, domain coverage, and benchmark support, as summarized in Table~\ref{tab:surveys}. 
Our survey stands out for offering a more comprehensive taxonomy, which provides a deeper and more nuanced understanding of the diverse PEFT methods, in contrast to the more limited or partial taxonomies found in existing reviews~\cite{ding2022delta,xu2023parameter,han2024parameter,wang2024parameter}. 
Furthermore, our survey is specifically tailored to the vision domain, whereas prior works predominantly focus on the NLP domain. This vision-centric emphasis addresses the unique challenges and developments in PEFT for visual tasks, which present distinct requirements compared to NLP applications. 
Finally, we introduce a benchmark for evaluating PEFT methods, setting our survey apart by providing practical tools for researchers to systematically compare and assess the performance of PEFT methods, which is not provided by previous reviews.

\begin{figure*}[t]
    \centering
    \vspace{-0.4cm}
    \includegraphics[width=0.92\linewidth]{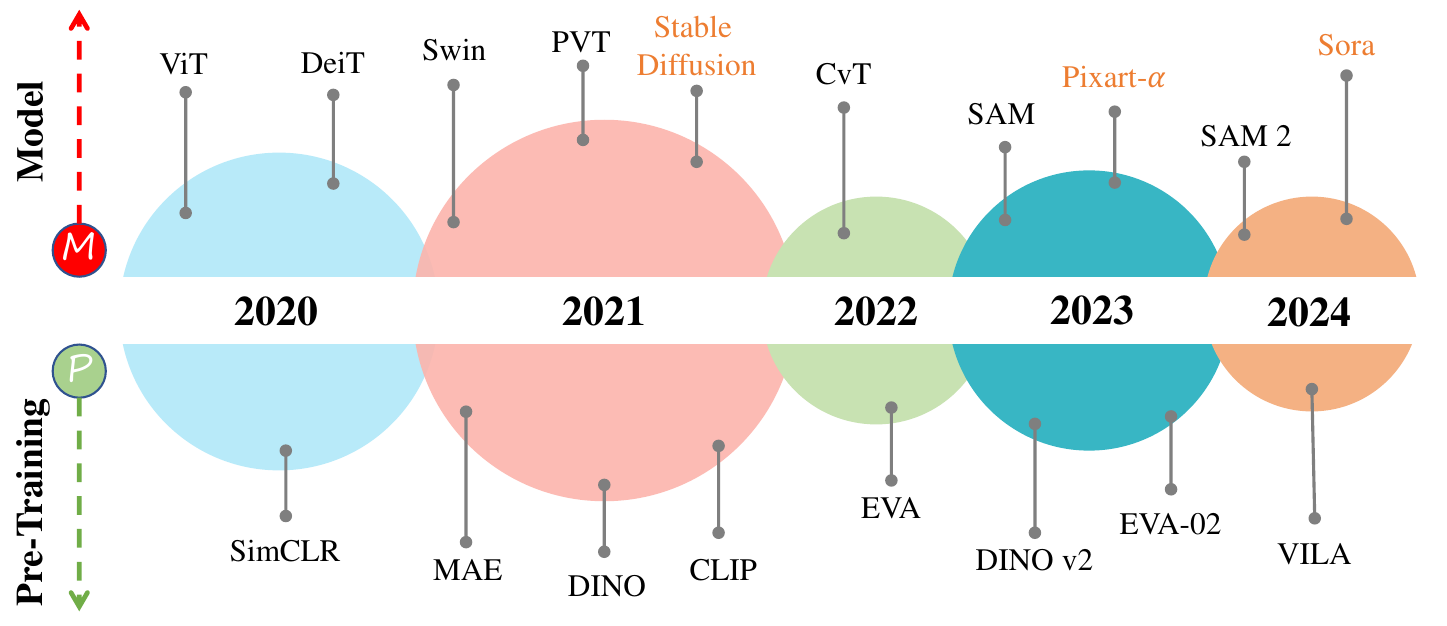}
    \vspace{-0.17in}
    \caption{
    \textbf{Representative Vision Foundation Models and Pre-Training Methods.} Our analysis primarily focuses on the significant advancements made between 2020 and 2024. Notably, models highlighted in \textcolor{orange}{orange} represent diffusion models.}
    \label{fig:models}
    \vspace{-0.15in}
\end{figure*}

\vspace{-0.15in}
\section{Preliminaries}
\label{sec:preliminary}
\subsection{Problem Definition}
\textnormal{Parameter-Efficient Fine-Tuning methods aim to adapt pre-trained models to new tasks by updating only a small number of parameters, rather than the entire model, thereby preventing overfitting while improving performance. Given a pre-trained model $M$ parametrized by 
    $\theta$, and a downstream task  $\mathcal{D} = \{(x_i, y_i)\}_{ i=1}^{|\mathcal{D}|}$, where $(x_i, y_i)$ serves as a ground-truth input-output pair of task $\mathcal{D}$,  parameter-efficient fine-tuning aims to adapt $\theta$ to task $\mathcal{D}$, where task-specific parameters increment $\Delta\theta$ is introduced with  $|\Delta\theta| \ll |\theta|$. The optimal parameters are found by optimizing the losses $\mathcal{L}$ on task $\mathcal{D}$:
    \begin{equation}
        \min_{\Delta\theta}\mathbb{E}_{(x_i, y_i) \in \mathcal{D}} \mathcal{L}(M_{\theta + \Delta\theta}(\hat{y_i}|x_i), y_i).
    \end{equation}
}

\vspace{-0.2in}
\subsection{Vision Transformer}
The vision transformer~\cite{dosovitskiy2020vit} consists of a patch embedding layer and $L$ transformer layers.
Given an image $x\in \mathbb{R}^{H\times W\times C}$, the patch embedding layer first splits and flattens the image $x$ into sequential patches $x_p\in \mathbb{R}^{N\times(P^2C)}$, where $(H, W)$ represents the height and width of the input image, $(P, P)$ is the resolution of each image patch, $C$ denotes the number of channels, and $N=HW/P^2$ is the number of image tokens. Then, $x_p$ is mapped to $x_0\in \mathbb{R}^{N\times d}$ with a trainable linear projection. The combination of a pre-appended $[cls]$ token and $x_0$ are the inputs of the transformer.

Each transformer layer consists of a multi-head attention (MHA) and a multilayer perceptron (MLP) module. In MHA, attention scores are computed using query ($Q$), key ($K$), and value ($V$) representations, along with projection matrices $W_{q}, W_{k}, W_{v} \in \mathbb{R}^{d\times d}$.  Given an input $x_{\ell-1}$ at the $\ell$-th layer, the attention is calculated as follows:
\vspace{-0.05cm}
\begin{align}
        Q = x_{\ell-1}W_{q},\ K = x_{\ell-1}W_{k},\ V = x_{\ell-1}W_{v}, \\
    \label{eq:attention}
    x_{\ell}^{\prime} = Attention(Q, K, V) = softmax(\frac{QK^{\top}}{\sqrt{d}})V.
\end{align}

The output tokens $x_{\ell}^{\prime}$ are further sent to a LayerNorm(LN) and an MLP block, which is formulated as follows:
\begin{equation}
    x_{\ell} = MLP(LN(x_{\ell}^{\prime})) + x_{\ell}^{\prime},
\end{equation}
where $x_{\ell}$ is the output of the $\ell$-th encoder layer.

Recent advances in vision transformer architectures have significantly enhanced performance in vision tasks. 
One line of work improves standard ViT by integrating additional or contextual information, with notable models such as Pyramid DeiT~\cite{touvron2021training} and Token to Token (T2T) ViT~\cite{yuan2021tokens}. Another line of work focuses on multi-scale ViTs using hierarchical designs to capture spatial details at varying scales, which is a capability limited in standard ViTs due to fixed token numbers and dimensions. 
The key models in this category include Pyramid ViT (PVT)~\cite{Wang2021PyramidVT} and Swin transformer~\cite{liu2021swin}.
A more comprehensive survey can be found in this literature~\cite{khan2022transformers}.

\vspace{-0.15in}
\subsection{Diffusion Model}
Diffusion model~\cite{NEURIPS2020_4c5bcfec} represents a class of generative models that learn to reverse a diffusion process. In this framework, the forward process progressively introduces noise to the data, while the reverse process attempts to reconstruct the original data from the noise.
Given an input image $x \in \mathbb{R}^{H \times W \times C}$, the forward diffusion process generates a sequence of noisy images $\{x_t\}_{t=1}^{\top}$, where $T$ denotes the total number of steps. 
At each step $t$, Gaussian noise is added to the image according to a variance schedule $\beta_t$, resulting in the noisy image $x_t$. The reverse process is modeled as the denoising neural network $\theta$, which takes a noisy image $x_t$ at step $t$ and predicts the mean and variance of the previous step $x_{t-1}$. The objective of training the diffusion model is to minimize the difference between the predicted and actual clean images. Specifically, the network learns to predict the noise added during the forward diffusion process. The model is trained to minimize the following loss:
\vspace{-0.15cm}
\begin{equation}
    \mathcal{L}_{\text{diffusion}} = \mathbb{E}_{t, x_0, \epsilon} \left[ \|\epsilon - \hat{\epsilon}_\theta(x_t, t)\|^2 \right],
\end{equation}
where $\epsilon$ is the noise added in the forward process, and $\hat{\epsilon}_\theta(x_t, t)$ is the predicted noise at step $t$.

In diffusion models, PEFT methods were initially applied to the Stable Diffusion (SD) series of models~\cite{rombach2022high,podell2023sdxl}, which integrate a Variational Autoencoder (VAE)~\cite{kingma2013auto} for more efficient latent space encoding and utilize a U-Net architecture~\cite{ronneberger2015u} as the denoising network. The denoising process in SD is guided by a noise schedule, and the model is often conditioned on additional input, such as text prompts, via cross-attention mechanisms. Beyond SD models, PEFT methods have also been incorporated into diffusion transformer (DiT) architectures~\cite{chen2023pixart,qin2025lumina}, which replace the U-Net architecture in SD with a transformer-based architecture for the denoising network. More details on the diffusion models can be found in~\cite{croitoru2023diffusion}.

\vspace{-0.2in}
\subsection{Pre-training} 
Pre-training methods are crucial for vision foundation models as they form the basis of their capabilities. 
These methods can be broadly categorized into supervised and self-supervised learning. Both paradigms have proven essential for training powerful vision models, with the latter showing greater promise for transferability and data efficiency in PEFT settings. 
To evaluate the effectiveness of a PEFT algorithm, it is generally necessary to test it on models that have undergone both types of pre-training.

\vspace{0.05cm}
\noindent \textbf{Supervised Pre-training.} These methods rely on large-scale annotated datasets and use classification objectives for pre-training. A notable example is ImageNet~\cite{deng2009imagenet}, which has served as a cornerstone for training high-performance vision models. 
Supervised pre-training excels in scenarios where labeled data is abundant and provides strong task-specific priors. Recently, the Segment Anything Model (SAM)~\cite{kirillov2023segment} demonstrated the power of dense, pixel-level supervision. Trained on a massive corpus of segmentation masks, SAM has achieved state-of-the-art performance in various segmentation tasks and illustrates the effectiveness of supervised pre-training in learning spatially rich representations. 
However, reliance on high-quality annotations remains a limiting factor in scaling such methods to broader domains.

\tikzstyle{my-box}=[
 rectangle,
 draw=hidden-draw,
 rounded corners,
 text opacity=1,
 minimum height=2em,
 minimum width=5em,
 inner sep=3pt,
 align=center,
 fill opacity=.5,
 ]
 \tikzstyle{leaf}=[my-box, minimum height=2em,
 fill=yellow!20, text=black, align=left,font=\scriptsize,
 inner xsep=2pt,
 inner ysep=4pt,
 ]

\begin{figure*}[t]
	\centering
    \vspace{-0.35cm}
	\resizebox{\textwidth}{!}{
		\begin{forest}
			forked edges,
			for tree={
				grow=east,
				reversed=true,
				anchor=base west,
				parent anchor=east,
				child anchor=west,
				base=left,
				font=\footnotesize,
				rectangle,
                fill=green!20,
				draw=hidden-draw,
				rounded corners,
				align=left,
				minimum width=4em,
				edge+={darkgray, line width=1pt},
				s sep=3pt,
				inner xsep=2pt,
				inner ysep=3pt,
				ver/.style={rotate=90, child anchor=north, parent anchor=south, anchor=center},
			},
			where level=1{text width=11em,font=\footnotesize, fill=orange!30}{},
			where level=2{text width=7.7em,font=\footnotesize, fill=purple!30}{},
			where level=3{text width=7.15em,font=\footnotesize, fill=pink!90}{},
			[
			PEFT Methods for PVMs, ver
			[
			Addition-based Tuning (\S\ref{sec:Addition-based}) 
			[
			  Adapter Tuning
			[
    			Adapter Design
    			[
                    AdaptFormer~{\cite{chen2022adaptformer},}
                    I2V-Adapter~{\cite{guo2024i2v},}\\
                    Convpass~{\cite{jie2022convolutional},}
                    PEA~{\cite{sharma2023lossless},}
                    T2I-Adapter~{\cite{mou2024t2i},}\\
                    ST-Adapter~{\cite{pan2022st},}
                    ArtAdapter~{\cite{chen2024artadapter},}
                    AIM~{\cite{yang2023aim},}\\
                    IP-Adapter~{\cite{ye2023ip-adapter},}
                    SimDA~{\cite{xing2024simda}}\\
    			, leaf, text width=14.5em
    			]
			]
                [
    			  Optimization
    			[
                    LoRand~{\cite{yin20231},}
                    SCT~{\cite{zhao2023sct},}
                    ARC~{\cite{dong2024efficient},}
                    SFA~{\cite{deng2023selective},}\\
                    MoSA~{\cite{zhang2023mosa},}
                    Opt-Adapter~{\cite{nowak2024towards},}
                    Mini~{\cite{marouf2024mini},}\\
                    Mem-Adapter~{\cite{xu2024memory},}
                    CoDA~{\cite{lei2023conditional}}   
    			, leaf, text width=14.5em
    			]
			]Taxonomy
			]
			[
			  Prompt Tuning
			[
    			Embedding Level
    			[
                    VPT~{\cite{jia2022visual},}   
                    CVP~{\cite{tsai2023convolutional},}
                    LPT~{\cite{dong2022lpt},}
                    E$^{2}$VPT~{\cite{han20232},}\\
                    ViPT~{\cite{zhu2023visual},}
                    IDPT~{\cite{zha2023instance},}
                    LION~{\cite{wang2023lion},}
                    DePT~{\cite{gao2022visual},} \\
                    SA$^{2}$VP~{\cite{pei2024sa2vp},}
                    EXPRES~{\cite{das2023learning},}
                    Fair-VPT~{\cite{park2024fair},}\\                    SHIP~{\cite{zhu2024semantic},}
                    G-VPT~{\cite{sohn2023visual},}
                    Pro-tuning~{\cite{nie2023pro}}
    			, leaf, text width=14.5em
    			]
			]
                [
    			  Pixel Level
    			[
                    VP~{\cite{bahng2022exploring},}
                    EVP-L~{\cite{liu2023explicit},}
                    P2P~{\cite{wang2022p2p},}
                    EVP~{\cite{wu2022unleashing},}\\
                    IML-VP~{\cite{chen2023understanding},}
                    ProSFDA~{\cite{hu2022prosfda},}
                    DAM-VP~{\cite{huang2023diversity}}
    			, leaf, text width=14.5em
    			]
			]
			]
                [
    			  Prefix Tuning
    			[
                    PATT~{\cite{yu2022towards},}
                    eTT~{\cite{xu2023exploring},}
                    LAE~{\cite{gao2023unified},}
                    VQT~{\cite{tu2023visual},}
                    Prefix-tuning~{\cite{li2021prefix},}
                    APT~{\cite{bandara2024attention},}\\
                    FedPerfix~{\cite{sun2023fedperfix}}
    			, leaf, text width=23.3em
    			]
			]
			[
			  Side Tuning
			[
    			Parameter Efficient
    			[
                    ControlNet~{\cite{zhang2023adding},}
                    SAN~{\cite{xu2023side},}
                    HST~{\cite{lin2023hierarchical},}\\
                    Uni-controlnet~{\cite{zhao2024uni},}
                    Side-Tuning~{\cite{zhang2020side},}\\
                    ViT-Adapter~{\cite{chen2022vision}}
    			, leaf, text width=14.5em
    			]
			]
                [
    			  Parameter \& \\Memory Efficient
    			[
                    SAM-LST~{\cite{chai2023ladder},}
                    E$^{3}$VA~{\cite{yin2023parameter},}
                    LoSA~{\cite{mercea2024time},}\\
                    LST~{\cite{sung2022lst},}
                    DTL~{\cite{fu2023dtl}}
    			, leaf, text width=14.5em
    			]
			]
			]
			]
			[
			  Partial-based Tuning (\S\ref{sec:Partial-based})
			[
                    Specification Tuning
                        [
                        Linear Probe~{\cite{kornblith2019better},}  
                        AdapterBias~{\cite{fu2022adapterbias},} 
                        DiffFit~{\cite{xie2023difffit}}
                        DP-BiTFiT~{\cite{bu2022differentially},}\\
                        LN-Tune~{\cite{basu2023strong},}
                        GPS~{\cite{zhang2024gradient},}
                        BitFit~{\cite{zaken2021bitfit},}
                        FedSelect~{\cite{tamirisa2024fedselect}}
                        , leaf, text width=23.3em
                        ]
			]
   			[
                    Reparameter Tuning
                        [
                            LoRA~{\cite{hu2021lora},} KronA~{\cite{edalati2022krona},} FacT~{\cite{jie2023fact},} EFFT~{\cite{chen2023aggregate},} 
                            SSF~{\cite{lian2022scaling},}
                            DnA~{\cite{jiang2022dna},}\\
                            Atten-Scale~{\cite{basu2023strong},} 
                            ALoRE~{\cite{du2024alore},}
                            KAdaptation~{\cite{he2023parameter},}
                            DMLoRA~{\cite{fang2025dropout},}\\
                            Concept sliders~{\cite{gandikota2024concept},}
                            RepAdapter~{\cite{luo2023towards}} 
                            ,leaf, text width=23.3em
                        ]
			]
			]
                [
			    Unified-based Tuning (\S\ref{sec:Unified-based})
                    [
                        Hybrid PEFT
            			[
                                V-PETL~{\cite{yu2022towards},}
                                NOAH~{\cite{liu2022neural},}
                                U-Tuning~{\cite{jiang2023rethinking},}
                                LAE~{\cite{gao2023unified},}
                                DAPT~{\cite{zhou2024dynamic},}\\
                                PETL-Vision~{\cite{mai2024lessons},}
                                DyT~{\cite{zhao2024dynamic}}
                                , leaf, text width=23.3em
            			]
                    ]
                    [
                        Cross-Tech \& PEFT \\Fusion
                        [
                                DyT~{\cite{zhao2024dynamic},}
                                Sparse Tuning~{\cite{liu2024sparse},}
                                FreqFit~{\cite{ly2024enhancing},}
                                HTA~{\cite{dongefficient},}
                                GIST~{\cite{ruan2024gist},}\\
                                Dyn-Adapter~{\cite{zhang2025dyn},}
                                SLS~{\cite{han2025straightforward}}
                                , leaf, text width=23.3em
            			]
                    ]
			]
                [
			    Multi-task Tuning (\S\ref{sec:multi-task})
        			[
                            Polyhistor~{\cite{liu2022polyhistor},}
                            VMT-Adapter~{\cite{xin2023vmt},} 
                            EMTAL~{\cite{zhongtransforming},}
                            MLoRE~{\cite{yang2024multi},}
                            VL-MPFT~{\cite{zhu2024vl},}
                            MmAP~{\cite{xin2024mmap},}\\
                            Hyperformer~{\cite{mahabadi2021parameter},}
                            TADFormer~{\cite{baek2025tadformer},}
                            MTLoRA~{\cite{agiza2024mtlora}}
                            , leaf, text width=32.7em
        			]
			]
			]
    \end{forest}
}

\caption{\textbf{Taxonomy of Parameter-Efficient Fine-Tuning Methods for Pre-Trained Vision Models.} Existing PEFT methods can be divided into 4 primary categories: \textbf{Addition-based Tuning}, which involves the integration of additional trainable neural modules or parameters into the PVMs; \textbf{Partial-based Tuning}, which focuses on selectively fine-tuning specific parameters within the original PVMs; \textbf{Unified-based Tuning}, which seeks to consolidate various PEFT approaches or incorporate other techniques; \textbf{Multi-task Tuning}, which emphasizes the synergy and complementary relationships between multiple tasks.} 
\vspace{-0.15in}
\label{PEFT_categorization_of_LLMs}
\end{figure*}
\vspace{0.05cm}
\noindent \textbf{Self-supervised Pre-training.}
Self-supervised learning (SSL) eliminates the need for manual labels and has become a mainstream strategy for scalable and generalizable pre-training. 
SSL in vision can be divided into two methodological families: 1) \textit{Contrastive learning methods}, which aim to learn invariant and discriminative representations by contrasting positive and negative sample pairs. 
Key image-based approaches include SimCLR~\cite{chen2020simple}, MoCo~\cite{he2020momentum}, and DINO~\cite{Caron2021EmergingPI}, which have achieved competitive results with their supervised counterparts. Multimodal extensions such as CLIP~\cite{Radford2021LearningTV}, ALIGN~\cite{Jia2021ScalingUV}, and VILA~\cite{lin2024vila} further align visual and textual representations to enable strong cross-modal understanding. 
To evaluate PEFT in a vision-specific context, this work focuses exclusively on the visual modules of these models, disregarding the text encoders and audio modalities. 2) \textit{Masked image modeling methods}, inspired by masked language modeling in NLP, have gained significant traction due to their ability to learn semantic structures through reconstruction. Methods such as MAE~\cite{he2022masked}, SimMIM~\cite{Xie2021SimMIMAS}, and EVA~\cite{Fang2022EVAET} mask image patches and train the model to predict missing content.

\begin{figure*}[t]
    \centering
    \vspace{-0.15in}
    \includegraphics[width=1.0\linewidth]{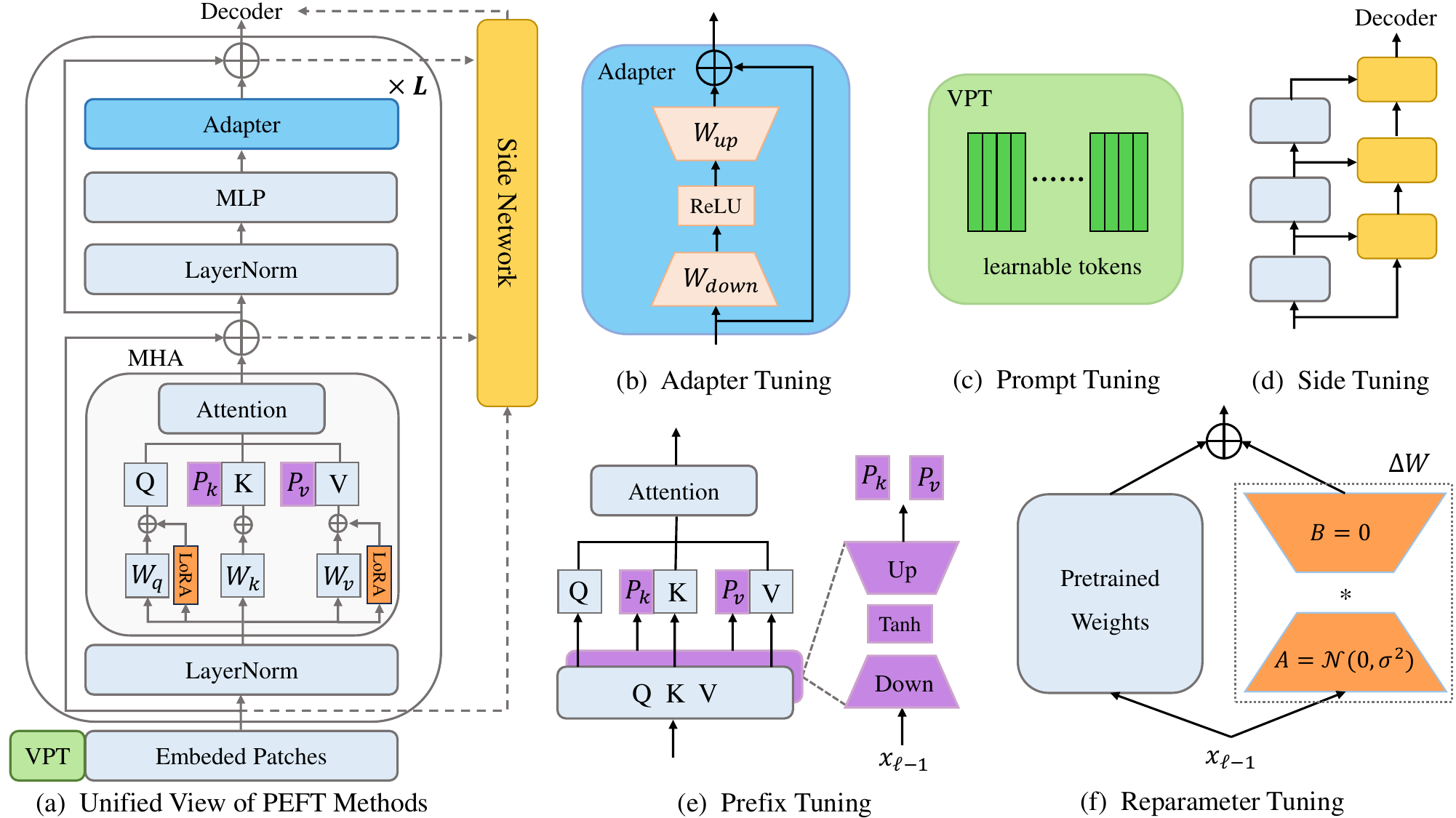}
    \vspace{-0.2in}
    \caption{
    \textbf{Detailed Architecture of PEFT Methods.} It covers the key components of different tuning strategies, including Adapter Tuning, Prompt Tuning, Prefix Tuning, Side Tuning, and Reparameter Tuning.}
    \vspace{-0.15in}
    \label{fig:method}
\end{figure*}
\vspace{-0.15in}
\section{Methodology}
\label{sec:taxonomy}
\subsection{Addition-based Tuning Methods}
\label{sec:Addition-based}
Addition-based tuning methods involve incorporating additional trainable modules or parameters into the original PVMs to learn task-specific information. This section discusses four primary branches of representative addition-based methods: adapter tuning, prompt tuning, prefix tuning, and side tuning.
\vspace{-0.1in}
\subsubsection{Adapter Tuning} 
The adapter is initially introduced in the NLP domain by~\cite{houlsby2019parameter} to achieve PEFT. Because of its remarkable effectiveness, it has also been successfully adopted in the CV field. This method integrates small neural modules, termed adapters, into the transformer layers. 
During the adaptation process, only these adapters are fine-tuned. The adapter architecture consists of a down-projection layer parameterized by $W_{down}\in \mathbb{R}^{d\times k}$ and an up-projection layer parameterized by $W_{up}\in \mathbb{R}^{k\times d}$, as shown in Figure~\ref{fig:method}{(b)}.
Here, $k$ (with $ k<<d$) serves to reduce the dimension of the representation to a lower rank. Furthermore, a ReLU layer is positioned between two layers to enable nonlinear projection. For a given input feature map $x_{\ell} \in \mathbb{R}^{N\times d}$, the adapter generates optimized features as follows:
\begin{equation}
\vspace{-0.2cm}
\hat{x_{\ell}}=\operatorname{ReLU}\left(x_{\ell}W_{down}\right) W_{up},
\end{equation}
where $W=[W_{down};W^{\top}_{up}] \in \mathbb{R}^{d \times 2k}$ denotes all the trainable parameters in the adapter. 

Adapter tuning methods in vision can be broadly divided into two main categories: 1) task-specific adapter architectures for various vision tasks and 2) advanced techniques to further reduce the trainable parameters in the adapter or to improve performance.

\noindent \textbf{Tasks-specific Adapters.} AdaptFormer~\cite{chen2022adaptformer} is a typical example. It represents the first instance of adapting vision transformers to a broad array of downstream visual recognition tasks using adapters. 
Unlike conventional approaches in NLP that utilize sequential insertion of adapters, AdaptFormer demonstrates that parallel insertion of adapters is more effective for vision tasks. 
Another example is Convpass~\cite{jie2022convolutional}, which identifies the limitations of current adapters due to their lack of strong inductive biases, thereby constraining their performance. 
To overcome this, Convpass incorporates trainable convolutional blocks into the adapter architecture, effectively merging the advantages of convolutional neural networks to enhance its capabilities. 
Furthermore, AIM~\cite{yang2023aim} introduces domain-specific adapters designed for spatial, temporal, and joint reasoning, while ST-Adapter~\cite{pan2022st} proposes a spatiotemporal adapter to improve a vision model's reasoning ability across spatial and temporal dimensions, catering to video understanding tasks. 
In the domain of robotic manipulation, Rob-Adapter~\cite{sharma2023lossless} uses the classic bottleneck architecture, commonly used in image classification, to achieve lossless adaptation for robotics-specific tasks. 
With the emergence of image and video generation tasks, adapter tuning methods have also been widely adopted in this field. 
These include their use for style embedding~\cite{mou2024t2i,chen2024artadapter}, model fine-tuning acceleration~\cite{xing2024simda}, and other applications that align with the unique demands of generative tasks.

\noindent \textbf{Reducing Trainable Parameters.}
Numerous methods focus on optimizing the adapter architecture to reduce the number of trainable parameters or improve performance. 
An example is LoRand~\cite{yin20231}, which constructs compact adapter structures through a low-rank synthesis approach. 
This technique reduces the parameters by representing both the down-projection layer $W_{down}$ and the up-projection layer $W_{up}$ as the product of three low-rank matrices. 
Similarly, ARC~\cite{dong2024efficient} introduces the idea of sharing the bottleneck operation's down- and up-projections across layers and employs low-dimensional re-composing coefficients to create layer-adaptive adapters. 
In order to unleash the potential of each parameter in the adapters, MoSA~\cite{zhang2023mosa} decomposes the standard adapter into multiple non-overlapping modules, stochastically activates these modules for sparse training, and subsequently merges them into a complete adapter post-tuning. 
Another innovative approach is SCT~\cite{zhao2023sct}, which adopts a selective channel tuning strategy. This method prioritizes tuning task-relevant channels, thereby significantly reducing parameter costs while maintaining performance. 
Furthermore, the insertion position of adapters within PVMs plays a critical role in their effectiveness. 
Although earlier studies~\cite{chen2022adaptformer} have explored this aspect experimentally, investigation has remained insufficient. 
Opt-Adapter~\cite{nowak2024towards} addresses this gap by providing a comprehensive analysis of the optimal insertion positions of the adapters, offering valuable insight to improve their integration into PVM.

\vspace{-0.2in}
\subsubsection{Prompt Tuning} 
Visual prompt tuning methods provide an alternative to injecting learnable modules into the transformer model. In such a method, the original input, whether it is an image embedding or the actual image, is wrapped with visual prompts. These prompts consist of additional trainable parameters or perturbations. They are uniquely adaptable parameters and can be optimized according to the specific task and the training data. The primary goal is to align the input distribution to the original pre-training data with task-specific prompts. 
Visual prompt tuning approaches can be categorized as: 1) inject a set of learnable parameters into the image embedding space, and 2) inject learnable perturbations around the border of the original input image.

\noindent \textbf{Learnable Parameters in Embedded Space.} 
VPT~\cite{jia2022visual} (including 
VPT-Shallow and VPT-Deep) is a
representative method (see Figure~\ref{fig:method}(c)).
VPT-Shallow integrates additional
$l$ learnable prompts, denoted as $P = [P_1], [P_2],...[P_l] \in \mathbb{R}^{l\times d}$, into the embeddings of the input patch $x_0\in \mathbb{R}^{N\times d}$. 
These prompts are then concatenated with path embedding to form the final input. This process can be expressed as follows:
\begin{equation}
    x_0 = concat(P, x_0) = [P, x_0] \in \mathbb{R}^{(l+N)\times d},
\end{equation}
where $[\cdot, \cdot]$ is the concatenation along the token dimension. VPT-Deep extends the capabilities of VPT-Shallow by introducing prompts into the input space of each transformer layer. During fine-tuning, only these prompts are updated, while all pre-trained parameters remain frozen. The computational cost of VPT-Deep is determined by the length of the prompt and the embedding dimension of the token, with experimental results demonstrating that longer prompts generally lead to improved performance. 
Building upon VPT, a series of subsequent works have adapted this method to specific tasks. 
DePT~\cite{gao2022visual} incorporates learnable visual prompts into vision transformers for data-efficient adaptation to the test time domain, and
CVP~\cite{tsai2023convolutional} introduces a self-supervised convolutional prompt to enhance robust visual perception.
In~\cite{dong2022lpt}, LPT focuses on optimizing shared prompts to extract generalizable features for long-tailed datasets. 
Recently, IDPT~\cite{zha2023instance} explores the application of visual prompt tuning on pre-trained point cloud models.
Beyond direct prompt integration, several schemes have proposed designing sub-networks to generate visual prompts dynamically. 
Pro-Tuning~\cite{nie2023pro} develops lightweight prompt blocks composed of three convolutional layers to produce task-specific discriminative prompts for individual downstream input images. 
In~\cite{wang2023lion}, LION introduces two implicit layers positioned at the beginning and end of the PVMs to serve as visual prompts, enriching both the input and the learned representations. 
On the other hand, 
ViPT~\cite{zhu2023visual} processes both RGB and auxiliary modality input through parallel patch embedding layers to generate corresponding RGB and prompt tokens. 
Despite the significant success of VPT, it still exhibits certain limitations. 
SA$^{2}$VP~\cite{pei2024sa2vp} shows standard prompt tuning methods to assume that all prompt tokens uniformly influence all image tokens, thereby lacking the capacity for fine-grained prompting. 
As such, SA$^{2}$ VP proposes the aligned and adapted visual prompt model, which introduces a two-dimensional prompt token map of equal (or scaled) size to the image token map. 
This design enables spatial alignment between the prompt map and the image token map, enhancing the model's ability to provide precise and spatially aware prompts.

\noindent \textbf{Learnable Perturbations in Images.}
Numerous methods focus on optimizing task-specific prompts at the pixel level, integrating these prompts directly with input images. 
VP~\cite{bahng2022exploring} introduces learnable perturbations around the image borders, eliminating the need for access to PVMs during the test phase. 
Building on this concept, EVP~\cite{wu2022unleashing} improves the approach by shrinking the input images, applying data augmentations, and padding the surrounding area with prompt information, enriching the input representation. 
DAM-VP~\cite{huang2023diversity} takes a divide-and-conquer approach by segmenting high-diversity datasets into smaller subsets and learning separate prompts for each subset. This strategy effectively addresses the challenges posed by the large diversity of data. 
Pixel-level prompt methods have demonstrated particular effectiveness in tasks such as image segmentation and point cloud processing.
EVP-L~\cite{liu2023explicit} leverages the high-frequency components of the input as prompts, specifically tailored for low-level structure segmentation tasks. 
On the other hand, ProSFDA~\cite{hu2022prosfda} introduces a zero-initialized learnable prompt directly into target images, improving performance in medical image segmentation. 
For point cloud analysis, P2P~\cite{wang2022p2p} innovatively converts point cloud data into colorful images, which are then used as visual prompts to adapt PVMs for various tasks.
In addition, ILM-VP~\cite{chen2023understanding} explores methods to better understand and improve the effectiveness of visual prompts. 
This method automatically remaps the source labels to the target labels, thereby improving the accuracy of target task.

\vspace{-0.1in}
\subsubsection{Prefix Tuning} 
Inspired by the success of prompt tuning, Prefix Tuning~\cite{li2021prefix} extends this paradigm by introducing learnable prefix matrices into the MHA mechanism of PVMs. 
Specifically, it incorporates two randomly initialized prefix matrices, $P_k, P_v \in \mathbb{R}^{l \times d}$, which are pre-pended to the keys and values within the MHA. 
This modification reformulates the attention computation in Equation~\ref{eq:attention} as follows:
\begin{equation}
Attention(Q, K, V) = softmax(\frac{Q[P_k, K]^{\top}}{\sqrt{d}})[P_v, V].
\end{equation}
However, random initialization can cause random noise, which affects the convergence of downstream fine-tuning tasks. 
To address this, PATT~\cite{yu2022towards} proposes a parallel attention mechanism to the original attention module without random initialization and employs two linear layers (parameterized by $W_{down}\in \mathbb{R}^{d\times k}$ and $W_{up}\in \mathbb{R}^{k\times l}$) and Tanh layers to transform prefix matrices (see Figure~\ref{fig:method}(e)).
Specifically, for the $\ell$-th transformer layer, given the output of the previous layer, ${x}_{\ell-1}$, a pair of prefix matrices is derived as follows:
\begin{equation}
    P_k, P_v = Tanh({x}_{\ell-1}W_{down})W_{up}.
\end{equation}

Building on a similar concept as PATT, FedPerfix~\cite{sun2023fedperfix} introduces a local adapter to generate prefixes and integrates them into the original self-attention layers. 
Subsequent studies have further extended the scope of prefix tuning.
For example, ETT~\cite{xu2023exploring} leverages advances in attentive prefix tuning (e.g., generating novel key-value pairs) for few-shot learning tasks, while LAM~\cite{gao2023unified} incorporates prefix tuning into a unified framework for continual learning.
Additionally, APT~\cite{bandara2024attention} applies prefix tuning to video action recognition tasks. 
On the other hand, VQT~\cite{tu2023visual} appends prefix vectors exclusively to the query $Q$, rather than increasing both the key $K$ and the value $V$.

\vspace{-0.1in}
\subsubsection{Side Tuning} 
In contrast to previous PEFT techniques that generally integrate extra modules or parameters straight into PVMs, side tuning takes an alternative approach by using a smaller, detached side network running alongside the PVMs, as depicted in Figure~\ref{fig:method}(d).

Early side tuning approaches focus on the design of the side network for parameter efficiency.
Side-Tuning~\cite{zhang2020side} employs a four-layer convolutional network as the additive side network, whose outputs are combined with representations of the PVMs in the final layer to address various tasks.
Expanding on this idea, SAN~\cite{xu2023side} introduces a two-branch side adapter network, one branch is responsible for mask proposal predictions, while the other computes attention biases applied to self-attention blocks for mask class recognition. 
ViT-Adapter~\cite{chen2022vision} incorporates a spatial prior module alongside two feature interaction operations, enabling the integration of image priors into the vision transformer (ViT) architecture without a complete redesign.
This design is particularly advantageous for dense prediction tasks, as it compensates for missing local information and enhances fine-grained, multi-scale feature organization. 
Similarly, HST~\cite{lin2023hierarchical} develops a hierarchical trainable side network that effectively utilizes the intermediate features of the pre-trained backbone to produce multi-scale features for predictions. 
Notably, ControlNet~\cite{zhang2023adding}, which has gained prominence in control image generation tasks, also exemplifies a side tuning method.

Beyond parameter efficiency, subsequent studies have highlighted that side tuning can achieve GPU memory efficiency through innovative designs. 
For example, LST~\cite{sung2022lst} separates trainable parameters from the backbone model by constructing a compact transformer-based side network. 
This separation eliminates the need for computationally expensive backpropagation through the large backbone, delivering substantial GPU memory savings. 
Building on LST, SAM-LST~\cite{chai2023ladder} incorporates an additional convolutional neural network as a complementary encoder within the SAM framework, achieving faster training and reduced resource consumption. 
However, since LST is not universally applicable to all PVMs, such as the Swin transformer, E$^3$VA~\cite{yin2023parameter} addresses this limitation by introducing a gradient backpropagation highway for low-rank adapters, ensuring compatibility with all PVMs while enhancing efficiency. 
Recently, DTL~\cite{fu2023dtl} designs a compact side network tailored specifically for ViTs, while LoSA~\cite{mercea2024time} develops a parallel side network comprising low-rank mixer modules to refine features extracted from a frozen backbone.

\vspace{-0.15in}
\subsection{Partial-based Tuning Methods}
\label{sec:Partial-based}
Partial-based tuning methods concentrate on adapting models by updating only a small subset of parameters. These approaches avoid modifying the internal structure of the model, ensuring efficiency and stability during the adaptation. This section delves into two strategies within this paradigm: specification tuning and reparameterization tuning.
\vspace{-0.1in}
\subsubsection{Specification Tuning} 
Specification tuning is an efficient approach that focuses on directly modifying a specific subset of parameters within PVMs, such as bias terms and LayerNorm parameters, which are critical for downstream tasks. 
By focusing on these key parameters and discarding those considered less relevant, this method achieves a balance between simplicity and effectiveness. Despite its straightforward nature, specification tuning has demonstrated remarkable performance. 
Linear Probe\cite{kornblith2019better} introduces a linear layer as a classifier on top of frozen PVMs in which all parameters of the PVMs remain unchanged, enabling an evaluation of the pre-training capabilities of the models. 
Linear Probe has been widely used in numerous PEFT methods. Furthermore, BitFit\cite{zaken2021bitfit} empirically shows that optimizing only the bias terms within a model is sufficient for effective fine-tuning. This can be represented as follows:
\begin{equation}
    x_{\ell} = x_{\ell-1}W_{\ell} + b_{\ell},
\end{equation}
where the weight parameters $W_{\ell}$ are kept frozen, and only the bias $b_{\ell}$ is optimized during the tuning process. 
This approach enables the model to effectively retain more 95\% of its performance on several benchmarks. 
Building on the principles of BitFit, DP-BiTFiT~\cite{bu2022differentially} combines the efficiency of the standard BiTFit approach to address downstream tasks involving sensitive data and achieves state-of-the-art accuracy for differentially private algorithms.
Similarly, DiffFit\cite{xie2023difffit} extends this idea by fine-tuning not only the bias terms but also newly introduced scaling factors in specific layers of diffusion models.
This strategy accelerates training and reduces model storage costs, making it particularly appealing for resource-constrained scenarios.
Meanwhile, AdapterBias~\cite{fu2022adapterbias} introduces a novel method that avoids modifying the existing bias terms in the PVMs. 
Instead, it focuses on the bias term in the MLP layers by incorporating a linear layer with a weight parameter $\alpha$ and a tunable vector $v$. This approach is expressed as:
\begin{equation}
    x_{\ell} = x_{\ell-1}W_{\ell} + b_{\ell} + \alpha \otimes v.
\end{equation}
Instead of tuning bias terms, LN-Tune~\cite{basu2023strong} introduces a robust PEFT baseline that fine-tunes only the LayerNorm parameters of the PVMs. 
In contrast to specification tuning methods, GPS~\cite{zhang2024gradient} introduces a novel concept by selecting the most relevant parameters for fine-tuning based on the gradient of the task. 
This gradient-based parameter selection ensures that only parameters closely related to the downstream task are updated. 

\vspace{-0.1in}
\subsubsection{Reparameterization Tuning} 
Reparameterization tuning methods introduce new learnable parameters during training, which are later integrated into the original PVMs by reparameterization during inference. 
This allows for efficient fine-tuning without permanently altering the original model structure. 
LoRA~\cite{hu2021lora} injects trainable low-rank matrices into transformer layers to approximate updates to the weights, as shown in Figure~\ref{fig:method}{(f)}. 
For a pre-trained weight matrix $W_{\ell}$, LoRA represents its update using a low-rank decomposition as follows:
\begin{equation}
    W_l^{'} = W_{\ell} + \bigtriangleup W = W_{\ell} + BA,
\end{equation}
where $B$ and $A$ are trainable parameters. Typically, LoRA modifies the query and value projection matrices in multi-head attention. 
LoRA has been widely used in various visual tasks, such as image generation~\cite{huang2024context,wu2024difflora}, object tracking~\cite{lin2024tracking}, and continual learning~\cite{wei2024online}, etc.
Furthermore, a growing body of follow-up research aims to refine and extend the capabilities of LoRA. KronA~\cite{edalati2022krona} shares structural similarities with LoRA, but replaces the low-rank decomposition with a Kronecker product decomposition, expressed as $\Delta W = B \otimes A$. 
This adjustment improves computational efficiency and significantly reduces the number of required floating-point operations (FLOPs). 
Building on these advancements, KAdaptation~\cite{he2023parameter} further generalizes weight updates by decomposing them into a sum of $n$ Kronecker products between shared slow weights $A_i$ and independent fast weights $B_i$. 
Additionally, it introduces a second-level decomposition by factorizing $B_i$ into the product of two low-rank matrices, $u_i$ and $v_i$:
\begin{equation}
    W + \bigtriangleup W = W + \sum\nolimits_{i=1}^{n} A_i\otimes B_i = \sum\nolimits_{i=1}^{n} A_i\otimes (u_iv_i^{\top}).
\end{equation}
Consequently, this reduction in trainable parameters significantly enhances the efficiency of model adaptation. 
Extending this concept further, FacT~\cite{jie2023fact} introduces a framework for tensorization-decomposition, where the weights of PVMs are tensorized into a unified 3D tensor and their increments are decomposed into lightweight factors. This innovative approach provides an efficient mechanism for storing weight updates, thereby offering a novel paradigm for managing PVM parameters.
Building on the foundation of FacT, EFFT~\cite{chen2023aggregate} focuses on minimizing redundancies both within and between layers, while maintaining computational efficiency without introducing additional latency. 

This method exemplifies the potential of tensor decomposition as a powerful tool to optimize model tuning processes. Beyond adapting pre-trained weight matrices, other works have explored alternative parameters within PVMs. 
For example, SSF~\cite{lian2022scaling} incorporates learnable scale and shift parameters to adjust feature representations, subsequently reparameterizing these adjustments in the MLP layer. 
RepAdapter~\cite{luo2023towards} highlights that adapter modules can be seamlessly integrated into PVMs through structural reparameterization, enabling efficient adaptation with zero inference overhead.

\begin{figure}[t]
   \begin{picture}(0,310)
     \put(15,170){\includegraphics[width=0.9\linewidth]{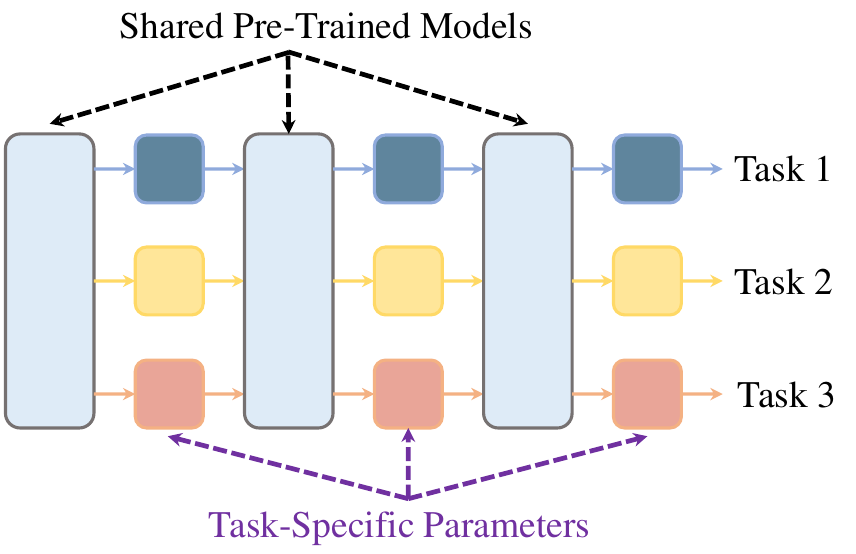}}
     \put(15,9){\includegraphics[width=0.9\linewidth]{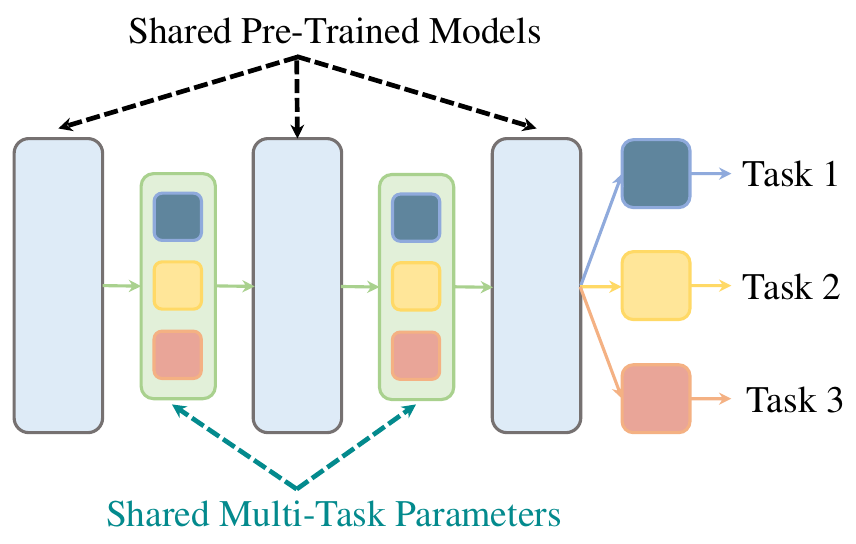}}
     \put(40,158){\small (a) Single Task Parameter-Efficient Fine-Tuning}
     \put(40,1){\small (b) Multi-Task Parameter-Efficient Fine-Tuning}
   \end{picture}
   \caption{\textbf{Comparison Between Single Task Tuning and Multi-Task Tuning.} (a) In single task tuning, each task is equipped with its own specific parameters, creating isolated and parallel execution paths for each task. This approach ensures task independence but lacks shared knowledge across tasks. (b) In multi-task tuning, tasks not only maintain their task-specific parameters but also leverage a set of shared parameters. This manner can facilitate the extraction of both task-shared and task-specific knowledge.}
   \vspace{-0.25in}
   \label{fig:multi-task-tuning}
\end{figure}

\vspace{-0.15in}
\subsection{Unified-based Tuning Methods}
\label{sec:Unified-based}
Unified-based tuning approaches provide a cohesive framework to consolidate diverse fine-tuning methods or integrate other techniques with PEFT methods. This unified paradigm simplifies the fine-tuning process while enhancing both its efficiency and overall effectiveness.

The effectiveness of a single PEFT method is often constrained by its fixed set of fine-tuning parameters. To overcome these limitations, several works have explored the combination of multiple PEFT methods to achieve performance improvements. For example, NOAH~\cite{liu2022neural} integrates Adapter, LoRA, and VPT into each transformer block and uses neural architecture search (NAS) to identify the optimal configurations for specific downstream tasks. This approach exemplifies a comprehensive strategy for fine-tuning optimization through the combination of multiple methods. LAM~\cite{gao2023unified} proposes a unified framework tailored for continual learning, offering flexibility by allowing any PEFT method to be reconfigured into a competitive approach for this setting. Similarly, V-PEFT~\cite{yu2022towards} provides a unified analysis of PEFT techniques, focusing on critical fine-tuning positions and providing a cohesive perspective on these approaches. 
On the other hand, U-Tuning~\cite{jiang2023rethinking} identifies a parallel structure underlying mainstream tuning methods, such as adapter, prefix, and prompt tuning, effectively reducing structural coupling in these approaches. 
%
%
Recently, PEFT-Vision~\cite{mai2024lessons} complements this by systematically analyzing existing PEFT techniques and offering practical recommendations on when and how to use them effectively. 

In addition to regular unified approaches, recent studies have incorporated strategies from other research domains into PEFT methods to simultaneously improve inference efficiency and reduce training parameters. For instance, DyT~\cite{zhao2024dynamic} introduces a token dispatcher for each transformer block, dynamically deciding whether tokens should be activated or deactivated. Activated tokens traverse both the entire block and an additional lightweight adapter module, while deactivated tokens bypass the block and are processed only by the adapter. Sparse Tuning~\cite{liu2024sparse} extends DyT by proposing a parameter-free token merge mechanism that retains semantically important tokens while condensing less relevant ones into representative tokens. It also employs dense adapters to integrate multi-level features from shallower layers, thereby enriching token representations. FreqFit~\cite{ly2024enhancing} introduces a frequency-based fine-tuning module that integrates seamlessly with existing PEFT methods, offering a simple yet effective way to improve model adaptation. 
In~\cite{dongefficient}, HTA leverages Householder matrices to construct Householder transformation-based adaptations, combining this technique with LoRA- and Adapter-based methods to enhance their efficiency and effectiveness. 
Recently, Dyn-Adapter~\cite{zhang2025dyn} proposes a dynamic architecture equipped with balanced early heads for multi-level feature extraction and introduces a bidirectional sparsity strategy to achieve robust generalization. 
This approach serves as a general efficiency booster for PEFT methods, further advancing their applicability and performance.

\setlength{\tabcolsep}{8pt} 
\renewcommand{\arraystretch}{1} 
\begin{table*}[t]
\renewcommand\arraystretch{1}
  \small
  \centering
  \vspace{-0.1in}
  \caption{\textbf{Comparison Between Different PEFT Methods.} We compare various characteristics, including NAM (No-Additional Modules), SP (Structure Preserving), IE (Inference Efficient), and ME (Memory Efficient).} 
  \vspace{-0.1in}
\scalebox{1}{
\footnotesize
    \begin{tabular}{cccccccc}
    \toprule
    \multirow{2}{*}{\textbf{Category}}
    & \multicolumn{4}{c}{\textbf{Characteristic}}        
    & \multirow{2}{*}{\makecell[c]{\textbf{Representative} \\\textbf{Algorithm}}} 
    & \multirow{2}{*}{\textbf{Year}}
    & \multirow{2}{*}{\textbf{ \#Trainable Params}} 
    \\ 
    \cmidrule{2-5}
    & \textbf{NAM} & \textbf{SP}  & \textbf{IE}  & \textbf{ME}   &  &\\  \midrule
    
    \cellcolor{cyan!8}{\textsc{Adapter Tuning}}&\cellcolor{cyan!8}\XSolid &\cellcolor{cyan!8}\XSolid &\cellcolor{cyan!8}\XSolid &\cellcolor{cyan!8}\XSolid &\cellcolor{cyan!8} AdaptFormer &\cellcolor{cyan!8}NeurIPS 2022 &\cellcolor{cyan!8}$ L \times (2dk)$ \\ 
    \midrule
    \textsc{Prompt Tuning} &\XSolid &\Checkmark &\XSolid &\XSolid & VPT-Deep &ECCV 2022 &  $L \times (ld)$\\ 
    \midrule
    \cellcolor{cyan!8}\textsc{Prefix Tuning}&\cellcolor{cyan!8}\XSolid &\cellcolor{cyan!8}\Checkmark &\cellcolor{cyan!8}\XSolid &\cellcolor{cyan!8}\XSolid & \cellcolor{cyan!8}PATT &\cellcolor{cyan!8}Arxiv 2022 &\cellcolor{cyan!8}$L \times 2 \times (dk + kl)$ \\ 
    \midrule
    \makecell[c]{\textsc{Side Tuning} } &\XSolid &\Checkmark &\XSolid &\Checkmark & \makecell[c]{LST} &NeurIPS 2022 &\#Params of subnetwork  \\
    \midrule
    \cellcolor{cyan!8}\textsc{Specification Tuning} &\cellcolor{cyan!8}\Checkmark &\cellcolor{cyan!8}\Checkmark &\cellcolor{cyan!8}\Checkmark &\cellcolor{cyan!8}\XSolid & \cellcolor{cyan!8}BitFit &\cellcolor{cyan!8}ACL 2022 &\cellcolor{cyan!8}$L \times (7 \times d)$  \\
    \midrule
    \makecell[c]{\textsc{Reparameter Tuning}} &\XSolid &\Checkmark &\Checkmark &\XSolid & \makecell[c]{LoRA} &ICLR 2022 &$L \times 2 \times (2dk)$\\
    \bottomrule
    \end{tabular}
    }
  \vspace{-0.15in}
  \label{tab:parameter}
\end{table*}

\vspace{-0.15in}
\subsection{Multi-task Tuning Methods}
\label{sec:multi-task}
In the field of visual PEFT, most existing methods focus on a single downstream task. 
Unlike single-task tuning, multi-task tuning necessitates the consideration of task interrelations and the exploitation of their complementarities to enhance the performance of each task. 
Consequently, multi-task tuning methods require acquiring task-shared and task-specific knowledge, as illustrated in Figure~\ref{fig:multi-task-tuning}. 
Polyhistor~\cite{liu2022polyhistor} addresses parameter-efficient multi-task tuning for vision tasks based on 
Hyperformer~\cite{mahabadi2021parameter}, a multi-task tuning approach for NLP.  
It achieves parameter efficiency in two key aspects: reducing the parameters within hyper-networks for multi-task architectures and decomposing weight matrix in task-specific adapters. 
On the other hand, VMT-Adapter~\cite{xin2023vmt} and MTDP~\cite{jiang2024task} incorporate task-conditional adapter designs to achieve multi-tasking capabilities. 
Similarly, MTLoRA~\cite{agiza2024mtlora} leverages a task-conditional LoRA structure to facilitate multi-task tuning. 
In~\cite{baek2025tadformer}, 
TADFormer introduces a parameter-efficient prompting mechanism tailored for task adaptation and proposes the Dynamic Task Filter (DTF), which captures task-specific information conditioned on the input context. 
Furthermore, MmAP~\cite{xin2024mmap} employs prompt-based designs to enhance multi-task tuning capabilities. Empirical experimental results from these studies demonstrate that multi-task tuning effectively harnesses task complementarities, significantly improving the performance of individual tasks.

\vspace{-0.15in}
\subsection{Discussion}
In the previous sections, we mainly categorized and summarized the various PEFT methods. However, a direct comparison between these methods has not yet been conducted. Therefore, in this section, we focus on analyzing and comparing the different PEFT methods in two key aspects: characteristic analysis and parameter analysis. Note that unified base tuning and multi-task tuning are not included in the comparison.

\vspace{-0.1in}
\subsubsection{Characteristic Analysis}
We summarize the characteristics of PEFT methods in Table~\ref{tab:parameter}: 
\begin{itemize}
    \item No-Additional Modules (NAM): Specification tuning is the only method that does not introduce any new modules, as it directly optimizes a subset of the parameters of the PVMs. In contrast, all other methods introduce additional modules or parameters to varying extents, which are leveraged during the fine-tuning.

    \item Structure Preserving (SP): Adapter tuning modifies the structure of PVMs by integrating additional adapter layers, thereby altering the model architecture. However, methods such as prompt tuning, prefix tuning, side tuning, and reparameter tuning preserve the original structure of PVMs, as they introduce new modules externally without affecting the core architecture. Specification tuning, which being a method that fine-tunes selected parameters of the PVMs, also preserves the model structure completely.

    \item Inference Efficient (IE): Introducing additional modules often increases the inference latency. However, re-parameter tuning is an exception, as it employs reparameterization techniques to integrate the additional modules into the original PVMs during inference, thereby mitigating the latency overhead. Other methods, such as prompt tuning or adapter tuning, may lead to slightly higher inference costs due to the presence of additional modules.

    \item Memory Efficient (ME): Side tuning stands out in terms of memory efficiency, as its gradient backpropagation process does not involve the PVMs. This decoupling reduces memory usage, making it particularly suitable for memory-constrained scenarios. Other methods, such as adapter tuning or prompt tuning, require additional memory due to the storage and optimization of the introduced modules.
\end{itemize}

\vspace{-0.15in}
\subsubsection{Parameter Analysis} 
To precisely compute the number of trainable parameters, we select a representative work from each taxonomy, as summarized in Table~\ref{tab:parameter}. In particular, BitFit exhibits the smallest number of trainable parameters since it updates only the bias terms within the PVMs. In contrast, LST demonstrates the largest number of trainable parameters due to its parallel subnetwork design, although it remains memory-efficient. This highlights the potential importance of optimizing subnetwork structures in future research. Furthermore, adaptFormer, PATT, and LoRA exhibit comparable parameter magnitudes, as they all integrate similar structural modifications into each transformer layer. Meanwhile, VPT-Deep maintains a slightly higher parameter count than BitFit. 
In practical applications, these methods require only 0.05\% to 10\% of the trainable parameters compared to full fine-tuning, yet they deliver comparable or even superior performance in downstream tasks.

\begin{table*}[!ht]
    \small
    \setlength{\tabcolsep}{6pt} 
    \renewcommand{\arraystretch}{1} 
    \vspace{-0.1cm}
    \caption{\textbf{Common Tasks and Datasets in the visual PEFT domain.} They can be categorized into three main tasks: image recognition (31 datasets), video recognition (7 datasets), and dense prediction (8 datasets). For each dataset, we analyze and record the number of classes, the size of the training, validation, and test datasets.}
    \scalebox{0.865}{%
        \begin{tabular}{lllllll}
        \toprule
        \textbf{Task} &	\textbf{Dataset}   & \textbf{Description}  & \textbf{\#Classes}    & \textbf{Train size} & \textbf{Val size}  & \textbf{Test size} \\ 
        \midrule 
             
          \multirow{34}{*}{\textbf{Image Recognition}} &\multicolumn{6}{c}{\textbf{Fine-Grained Visual Classification (FGVC)}~\cite{jia2022visual}}\\
 
        \cmidrule{2-7}
        & \cellcolor{cyan!8}CUB-200-2011~\cite{wah2011caltech} & \cellcolor{cyan!8}Fine-grained bird species recognition &\cellcolor{cyan!8}200 &\cellcolor{cyan!8}5,394	&\cellcolor{cyan!8}600 &\cellcolor{cyan!8}5,794	\\

          &NABirds~\cite{van2015building} &Fine-grained bird species recognition &55 &21,536	&2,393	&24,633	\\

        & \cellcolor{cyan!8}Oxford Flowers~\cite{nilsback2008automated} & \cellcolor{cyan!8}Fine-grained flower species recognition &\cellcolor{cyan!8}102 &\cellcolor{cyan!8}1,020	&\cellcolor{cyan!8}1,020	&\cellcolor{cyan!8}6,149  \\

        &Stanford Dogs~\cite{dataset2011novel}   &Fine-grained dog species recognition  &120  &10,800	&1,200 &8,580 \\

        & \cellcolor{cyan!8}Stanford Cars~\cite{gebru2017fine} & \cellcolor{cyan!8}Fine-grained car classification  &\cellcolor{cyan!8}196   &\cellcolor{cyan!8}7,329	&\cellcolor{cyan!8}815	&\cellcolor{cyan!8}8,041 \\
        \cmidrule{2-7}\
        
        & \multicolumn{6}{c}{\textbf{Visual Task Adaptation Benchmark (VTAB-1k)}~\cite{zhai2019large}} \\
        \cmidrule{2-7}

        &\cellcolor{cyan!8}CIFAR-100~\cite{krizhevsky2009learning} &\cellcolor{cyan!8}\multirow{7}{5cm}{ } &\cellcolor{cyan!8}100 &\cellcolor{cyan!8} \multirow{7}{*}{} &\cellcolor{cyan!8}\multirow{7}{*}{} &\cellcolor{cyan!8}10,000 \\

        & Caltech101~\cite{fei2006one} & &102 & & &6,084  \\

        &	\cellcolor{cyan!8}DTD~\cite{cimpoi2014describing} &\cellcolor{cyan!8} &\cellcolor{cyan!8}47 &\cellcolor{cyan!8}&\cellcolor{cyan!8}&\cellcolor{cyan!8}1,880  \\

        &	Flowers102~\cite{nilsback2008automated} &\textbf{\textit{Natural}}-tasks that contain natural &102 &800 &200 &6,149	\\

        &	Pets~\cite{parkhi2012cats} &images captured using standard cameras. &37 &&&3,669	\\

        &   \cellcolor{cyan!8}SVHN~\cite{netzer2011reading} &\cellcolor{cyan!8} &\cellcolor{cyan!8}10 &\cellcolor{cyan!8}&\cellcolor{cyan!8}&\cellcolor{cyan!8}26,032	\\

        &	Sun397~\cite{xiao2010sun} &&397 &&&21,750	\\
        \cmidrule{2-7}

        &   \cellcolor{cyan!8}Patch Camelyon~\cite{veeling2018rotation} &\cellcolor{cyan!8} &\cellcolor{cyan!8}2 &\cellcolor{cyan!8}\multirow{4}{*}{800} &\cellcolor{cyan!8}\multirow{4}{*}{200} &\cellcolor{cyan!8}32,768 	\\
        
        &  EuroSAT~\cite{helber2019eurosat} &\textbf{\textit{Specialized}}-tasks with images from &10 &  & &5,400		\\

        &	Resisc45~\cite{cheng2017remote} &specialized equipment (e.g., medical). &45 && &6,300		\\

        &	\cellcolor{cyan!8}Retinopathy~\cite{graham2015kaggle}  &\cellcolor{cyan!8} &\cellcolor{cyan!8}5 &\cellcolor{cyan!8}&\cellcolor{cyan!8}&\cellcolor{cyan!8}42,670		\\
        \cmidrule{2-7}
            
        &	Clevr/count~\cite{johnson2017clevr}  &\multirow{8}{5.13cm}{} &8	&\multirow{8}{*}{} &\multirow{8}{*}{} &15,000 \\
        
        &	\cellcolor{cyan!8}Clevr/distance~\cite{johnson2017clevr}  &\cellcolor{cyan!8}&\cellcolor{cyan!8}6 &\cellcolor{cyan!8}&\cellcolor{cyan!8}&\cellcolor{cyan!8}15,000	\\
        
        &	DMLab~\cite{beattie2016deepmind}  &\textbf{\textit{Structured}}-geometric reasoning&6 &&&22,735			\\	
        
        &	KITTI/distance~\cite{geiger2013vision} &tasks (e.g., object counting). &4 &800 &200 &711		\\  
        
        &	\cellcolor{cyan!8}dSprites/location~\cite{dsprites17} &\cellcolor{cyan!8} &\cellcolor{cyan!8}16 &\cellcolor{cyan!8} &\cellcolor{cyan!8} &\cellcolor{cyan!8}73,728	\\
        
        &	dSprites/orientation~\cite{dsprites17} &&16 &&&73,728	\\
            
        &	\cellcolor{cyan!8}SmallNORB/azimuth~\cite{lecun2004learning} &\cellcolor{cyan!8} &\cellcolor{cyan!8}18 &\cellcolor{cyan!8}&\cellcolor{cyan!8} &\cellcolor{cyan!8}12,150	\\

        &	SmallNORB/elevation~\cite{lecun2004learning}  &&9 &&&12,150	\\

        \cmidrule{2-7}
        &	\multicolumn{6}{c}{\textbf{General Image Recognition Datasets}} \\
        \cmidrule{2-7}

        & \cellcolor{cyan!8}CIFAR-10~\cite{krizhevsky2009learning} &\cellcolor{cyan!8}&\cellcolor{cyan!8}10 & \cellcolor{cyan!8}50,000 &\cellcolor{cyan!8}- &\cellcolor{cyan!8}10,000 \\

        & CIFAR-100~\cite{krizhevsky2009learning} &\multirow{1}{*}{General image recognition} &100 & 50,000 &- &10,000 \\
        
        & \cellcolor{cyan!8}ImageNet-1k~\cite{deng2009imagenet} &\cellcolor{cyan!8} & \cellcolor{cyan!8}1,000 &\cellcolor{cyan!8}1,281,167 & \cellcolor{cyan!8}50,000 &\cellcolor{cyan!8}50,000 \\

        \cmidrule{2-7}
        &	\multicolumn{6}{c}{\textbf{Domain Generalization Datasets}} \\
        \cmidrule{2-7}
        &ImageNet-V2~\cite{recht2019imagenet} &\multirow{4}{*}{} &1,000 &- &- &10,000\\
        &\cellcolor{cyan!8}ImageNet-Sketch~\cite{wang2019learning} &\cellcolor{cyan!8} &\cellcolor{cyan!8}1,000 &\cellcolor{cyan!8}- &\cellcolor{cyan!8}- &\cellcolor{cyan!8}50,889 \\
        &ImageNet-A~\cite{hendrycks2021natural} &Domain generalization recognition &200 &- &- &7,500 \\
        &\cellcolor{cyan!8}ImageNet-R~\cite{hendrycks2021many} &\cellcolor{cyan!8}&\cellcolor{cyan!8}200 &\cellcolor{cyan!8}- &\cellcolor{cyan!8}- &\cellcolor{cyan!8}30,000 \\

        \midrule
        \multirow{7}{*}{\textbf{Video Recognition}}
        &Kinetics-400~\cite{kay2017kinetics}   & \multirow{5}{*}{} &400 &240,436 &- &19,787\\

        &\cellcolor{cyan!8}Kinetics-700~\cite{carreira2019short} &\cellcolor{cyan!8}&\cellcolor{cyan!8}700 &\cellcolor{cyan!8}545,000 &\cellcolor{cyan!8}35,000 &\cellcolor{cyan!8}70,000\\
        
        &Something-Something-v2~\cite{goyal2017something} &Human video action recognition &174 &168,913 &24,777 &27,157 \\
        
        &\cellcolor{cyan!8}HMDB51~\cite{kuehne2011hmdb} &\cellcolor{cyan!8}&\cellcolor{cyan!8}51 &\cellcolor{cyan!8}3,500 &\cellcolor{cyan!8}1,500 &\cellcolor{cyan!8}1,849 \\

        &UCF-101~\cite{soomro2012ucf101} & &101 &9,537 &- &3,783 \\
        \cmidrule{2-7}
        &\cellcolor{cyan!8}Diving-48~\cite{li2018resound} &\cellcolor{cyan!8}Diving video action recognition &\cellcolor{cyan!8}48 &\cellcolor{cyan!8}15,900 &\cellcolor{cyan!8}- &\cellcolor{cyan!8}2,000 \\

        &EPIC-KITCHENS-100~\cite{damen2022rescaling} & Fine-grained video action recognition &100 &67,217 &9,668 &13,092\\
                
        \midrule
        \multirow{11}{*}{\textbf{Dense Prediction}}
        &\cellcolor{cyan!8}MS COCO~\cite{lin2014microsoft} &\cellcolor{cyan!8}&\cellcolor{cyan!8}80 &\cellcolor{cyan!8}118,000 &\cellcolor{cyan!8}- &\cellcolor{cyan!8}5,000\\
        &ADE20K~\cite{zhou2019semantic} &Nature image segmentation &150 &20,210 &- &2,000 \\
        &\cellcolor{cyan!8}PASCAL VOC~\cite{everingham2015pascal} &\cellcolor{cyan!8}&\cellcolor{cyan!8}21 &\cellcolor{cyan!8}1,464 &\cellcolor{cyan!8}- &\cellcolor{cyan!8}1,449 \\

        \cmidrule{2-7}
        &ADOM~\cite{ma2021abdomenct} & &7 &20 &5 &15 \\
        &\cellcolor{cyan!8}SPLEN~\cite{antonelli2022medical} &\cellcolor{cyan!8} &\cellcolor{cyan!8}2 &\cellcolor{cyan!8}41 &\cellcolor{cyan!8}9 &\cellcolor{cyan!8}50\\
        &MOMO~\cite{antonelli2022medical} & Medical image segmentation &2 &200	&50	&50 \\
        &\cellcolor{cyan!8}BRAST~\cite{al2020dataset} &\cellcolor{cyan!8} &\cellcolor{cyan!8}4 &\cellcolor{cyan!8}285 &\cellcolor{cyan!8}66 &\cellcolor{cyan!8}191 \\
        &SEGRAP~\cite{luo2025segrap2023} &&3 &300 &100	&100 \\
	\bottomrule
	\end{tabular}
    }
    \label{table: datasets}
    \vspace{-0.12in}
\end{table*}

\vspace{-0.15in}
\section{Tasks and Datasets}
\label{task_dataset}
In this section, we provide an overview of the popular tasks and datasets within the visual PEFT domain, as summarized in Table~\ref{table: datasets}. The tasks can be categorized into three primary groups: image recognition, video recognition, and dense prediction. Typically, each task is associated with 7 to 31 datasets, all of which are distributed under permissive licenses that facilitate their use for research purposes. In addition, PEFT has a wide range of applications in image and video generation tasks, which we also elaborate on.
\vspace{-0.15in}
\subsection{Image Recognition Task}
Image recognition represents a foundational application within the visual PEFT domain. A total of 31 commonly used image recognition datasets are summarized in Table~\ref{table: datasets}. These datasets can be broadly grouped into four categories.
\vspace{-0.1in}
\subsubsection{Fine-Grained Visual Classification (FGVC)}
The FGVC category includes 5 fine-grained visual classification datasets: CUB-200-2011~\cite{wah2011caltech}, NABirds~\cite{van2015building}, Oxford Flowers~\cite{nilsback2008automated}, Stanford Dogs~\cite{dataset2011novel}, and Stanford Cars~\cite{gebru2017fine}. 
VPT~\cite{jia2022visual} incorporates FGVC datasets into the PEFT domain, establishing these datasets as pivotal benchmarks in this area. In particular, VPT applies preprocessing to the original datasets. For datasets where only training and testing splits are publicly available, VPT typically partitions 90\% of the training set for training purposes while reserving the remaining 10\% for validation. This validation set is subsequently utilized for hyperparameter selection.

\vspace{-0.1in}
\subsubsection{Visual Task Adaptation Benchmark (VTAB)}
VTAB~\cite{zhai2019large} consists of 19 diverse visual classification datasets, organized into three distinct domains: 1) \textbf{\textit{Natural}} - datasets containing natural images captured using standard cameras. This group includes Caltech101~\cite{fei2006one}, CIFAR100~\cite{krizhevsky2009learning}, DTD~\cite{cimpoi2014describing}, Flowers102~\cite{nilsback2008automated}, Pets~\cite{parkhi2012cats}, Sun397~\cite{xiao2010sun}, and SVHN~\cite{netzer2011reading}; 2) \textbf{\textit{Specialized}} - datasets comprising images acquired using specialized equipment, such as medical imaging and satellite imagery. 
This group includes Resisc45~\cite{cheng2017remote}, EuroSAT~\cite{helber2019eurosat}, Patch Camelyon~\cite{veeling2018rotation}, and Diabetic Retinopathy~\cite{graham2015kaggle}; 3) \textbf{\textit{Structured}} - datasets requiring geometric reasoning, such as object counting. This group includes Clevr~\cite{johnson2017clevr}, dSprites~\cite{dsprites17}, SmallNORB~\cite{lecun2004learning}, DMLab~\cite{beattie2016deepmind}, and KITTI~\cite{geiger2013vision}. Each dataset within VTAB is limited to 1000 training examples.

\vspace{-0.1in}
\subsubsection{General Image Recognition Datasets}
General image recognition tasks are also extensively studied in the visual PEFT domain, with CIFAR-100~\cite{krizhevsky2009learning} and ImageNet-1K~\cite{deng2009imagenet} being among the most commonly used datasets. CIFAR-100 consists of 60,000 images distributed across 100 categories. In contrast, ImageNet-1K is a significantly larger dataset, comprising 1.28M training images and 50,000 validation images spanning 1,000 categories. These large-scale datasets serve as critical benchmarks for advancing object recognition in the field.
\vspace{-0.1in}
\subsubsection{Domain Generalization Image Recognition Datasets}
Since domain shift is prevalent in real-world applications, several PEFT studies~\cite{liu2022neural,lian2022scaling} have focused on evaluating the domain generalization capabilities of their proposed methods. These studies typically train models on ImageNet and then directly evaluate them on four ImageNet variants that exhibit different types of domain shifts. 
 i) ImageNet-V2~\cite{recht2019imagenet}, which is collected from sources different from those of ImageNet but adheres to the same collection protocol; ii) ImageNet-Sketch~\cite{wang2019learning}, comprising sketch images corresponding to the same 1,000 classes as ImageNet; iii) ImageNet-A~\cite{hendrycks2021natural}, which contains adversarially filtered images designed to challenge standard classifiers; and iv) ImageNet-R~\cite{hendrycks2021many}, a rendition dataset that includes artistic and styled variations of ImageNet images.

\vspace{-0.12in}
\subsection{Video Action Recognition Task}
Seven public tasks are widely adopted in action-recognition research, each emphasizing a distinct aspect of action understanding.
Kinetics-400~\cite{kay2017kinetics} (240\,K/20\,K 10-s clips, 400 classes) is the de-facto pre-training corpus; its larger sibling Kinetics-700~\cite{carreira2019short} (more than 600\,K clips, 700 classes) stresses class-diversity shift. 
Something-Something-v2~\cite{goyal2017something} (170\,K/50\,K 2–6 s hand–object clips, 174 templated labels) demands fine-grained temporal reasoning, whereas HMDB51~\cite{kuehne2011hmdb} (6,766 clips, 51 actions) and UCF-101~\cite{soomro2012ucf101} (13,320 clips, 101 actions) remain popular mid-scale “in-the-wild’’ tests of appearance robustness. 
For fine-pose discrimination, we include Diving-48~\cite{li2018resound} (18,000 professional diving clips, 48 categories), and for first-person long-horizon interaction, we use EPIC-KITCHENS-100~\cite{damen2022rescaling} (\(\sim\)100 h egocentric kitchen video, more than 700 action labels). 
Together, these tasks cover increasing class diversity, temporal complexity, viewpoint variation, subtle pose changes, and egocentric hand–object interactions.
\vspace{-0.2in}
\subsection{Dense Prediction Task}
In dense prediction task, PEFT has been effectively utilized for both natural image segmentation and medical image segmentation. 
\vspace{-0.1in}
\subsubsection{Natural image segmentation}
Natural-image segmentation has emerged as a principal test-bed for dense-prediction PEFT, and three reference datasets now capture its complementary challenges.
MS-COCO~\cite{lin2014microsoft} (80 everyday categories with instance masks) focuses on instance-level localisation in cluttered scenes; 
ADE20K~\cite{zhou2017scene} (\(\sim\)20,000 indoor/outdoor images with pixel labels for 150 classes) targets full-scene parsing; 
PASCAL VOC~\cite{everingham2015pascal} (pixel annotations for 20 canonical categories) offers a compact test-bed for rapid ablation and cross-dataset transfer. 
Collectively, these datasets span instance vs.\ semantic segmentation, varied scene complexity, and a range of annotation scales.
\vspace{-0.1in}
\subsubsection{Medical Segmentation Datasets} 
In addition to the common dense prediction datasets,~\cite{peng2024parameter} also expolored medical image segmentation datasets, including ADOM~\cite{ma2021abdomenct}, SPLEN~\cite{antonelli2022medical}, MOMO~\cite{antonelli2022medical}, BRAST~\cite{al2020dataset} and SEGRAP~\cite{luo2025segrap2023}. These datasets encompass a variety of imaging modalities, such as CT, MRI, X-ray, and fundus images, and address 2D and 3D segmentation tasks. They feature a diverse range of anatomical structures, including abdominal organs, brain tumors, and retinal features, enabling applications in fields such as radiology, oncology, and ophthalmology. These datasets represent significant advances in medical image analysis, contributing to improved diagnostic precision and more effective treatment planning.

\begin{figure*}[t]
    \centering
    \vspace{-0.13in}
    \includegraphics[width=1.0\linewidth]{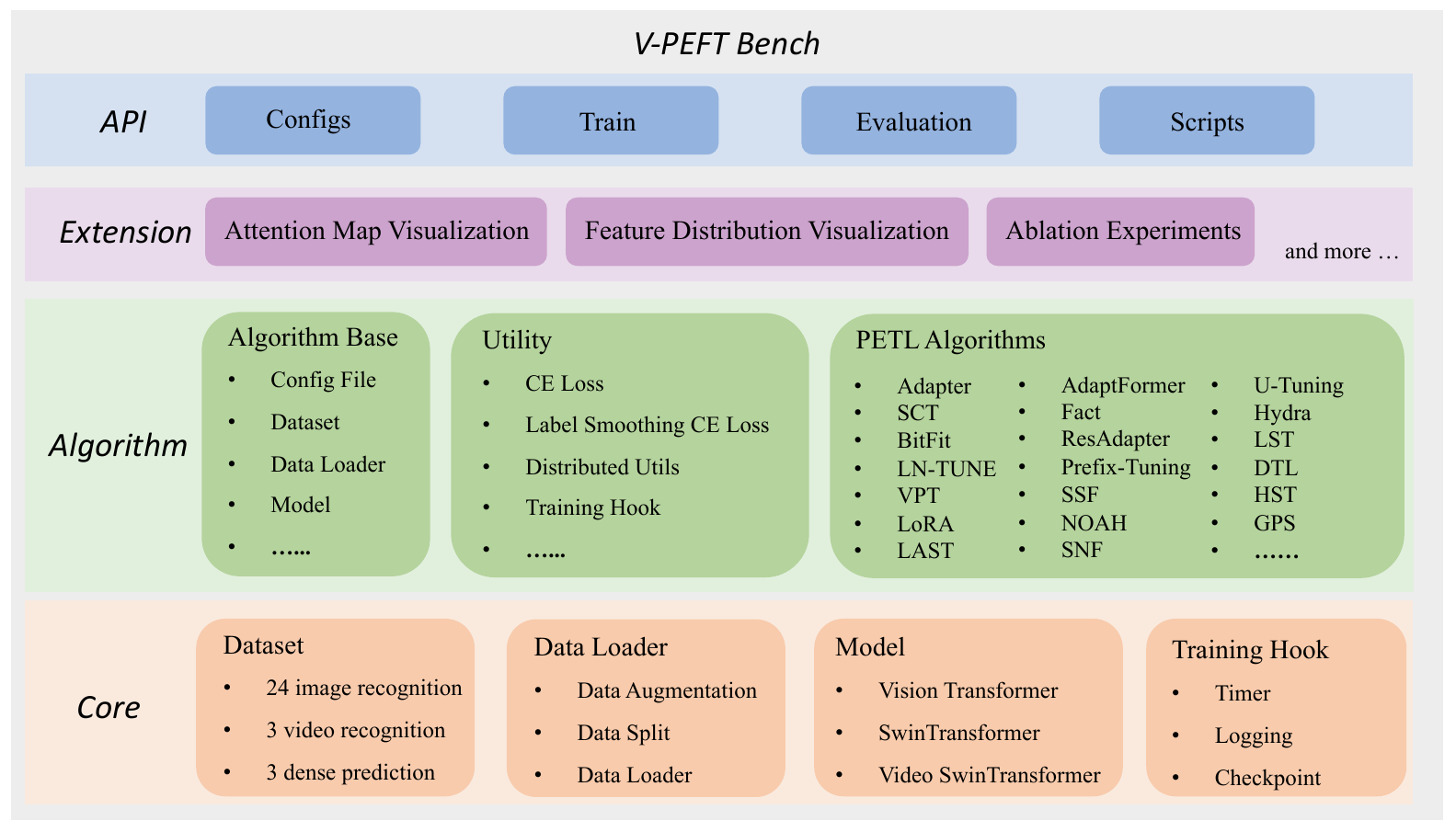}
    \vspace{-0.25in}
    \caption{
    \textbf{Overview of the V-PEFT Bench Codebase Architecture.} 
    \textcolor{black}{The Codebase comprises four modular layers: \textbf{Core}, which includes datasets, data loaders, model architectures (e.g., ViT, Swin), and training hooks; \textbf{Algorithm}, covering base modules, PETL methods (e.g., Adapter, LoRA, Prefix-Tuning), and utilities such as loss functions and hooks; \textbf{Extension}, offering tools for attention map visualization and feature distribution analysis; and \textbf{API}, which supports configuration, training, evaluation, and scripting. This architecture enables flexible experimentation and seamless integration of PEFT techniques.
    }}
    \label{fig:structure}
    \vspace{-0.2in}
\end{figure*}

\vspace{-0.15in}
\section{Benchmark}
\label{sec:benchmark}
In this section, we introduce the \textbf{Unified \underline{V}isual \underline{P}arameter-\underline{E}fficient \underline{F}ine-\underline{T}uning \underline{Bench}mark} (\textbf{V-PEFT Bench}), a comprehensive visual PEFT benchmark that spans 19 image recognition datasets (VTAB-1k), 2 video action recognition datasets (SSv2 and HMDB51), and 3 dense prediction datasets (MS COCO, ADE20K, and PASCAL VOC), as detailed in Section~\ref{task_dataset}. 
V-PEFT Bench facilitates consistent evaluation across a wide range of vision tasks by providing extensive assessments of 25 PEFT algorithms, with hyperparameter optimization conducted on various pre-trained vision transformers. 
Given the absence of a unified evaluation metric in the PEFT field that holistically accounts for both task performance and the number of trainable parameters, we propose the \textbf{\underline{P}erformance-\underline{P}arameter \underline{T}rade-off} (\textbf{PPT}) metric, which facilitates direct comparison of algorithms through one single integrated metric. 
Furthermore, the V-PEFT Bench includes t-SNE and attention map visualizations to offer deeper insights into the behavior and effectiveness of PEFT algorithms.

\vspace{-0.15in}
\subsection{Codebase Structure of V-PEFT Bench}
We first provide an overview of the codebase structure of the V-PEFT Bench, which is organized into four abstract layers, as shown in Figure~\ref{fig:structure}.

\renewcommand{\arraystretch}{1} 
\begin{table*}[t]
	\setlength\tabcolsep{2pt}
	\centering
    \vspace{-0.1in}
    \caption{\textbf{Benchmark Results on VTAB.} We evaluate 2 conventional fine-tuning algorithms and 18 PEFT algorithms on 19 datasets with ViT-B/16 models pre-trained on ImageNet-21K. We highlight the \textcolor{red}{best} and the \textcolor{citecolor}{second} results.}
    \vspace{-0.12in}
	\resizebox{\textwidth}{!}{
        \begin{tabular}{ccccccccccccccccccccccc}
			\toprule

            \multirow{8}{*}{\textbf{Method}} & 
            \multicolumn{6}{c}{\textbf{Natural}} & 
            \multicolumn{4}{c}{\textbf{Specialized}} & 
            \multicolumn{8}{c}{\textbf{Structured}} \\
            \cmidrule(lr){2-7}
            \cmidrule(lr){8-11}
            \cmidrule(lr){12-20}

            \\
			& \rotatebox{90}{\textbf{CIFAR-100}} & \rotatebox{90}{\textbf{Caltech101}} & \rotatebox{90}{\textbf{DTD}} & \rotatebox{90}{\textbf{Flowers102}} & \rotatebox{90}{\textbf{Pets}} & \rotatebox{90}{\textbf{SVHN}}  & \rotatebox{90}{\textbf{Sun397}} & \rotatebox{90}{\textbf{Patch Camelyon}~} & \rotatebox{90}{\textbf{EuroSAT}}   & \rotatebox{90}{\textbf{Resisc45}}  & \rotatebox{90}{\textbf{Retinopathy}} & \rotatebox{90}{\textbf{Clevr/count}} & \rotatebox{90}{\textbf{Clevr/distance}}  & \rotatebox{90}{\textbf{DMLab}} & \rotatebox{90}{\textbf{KITTI/distance}}  & \rotatebox{90}{\textbf{dSprites/loc}} & \rotatebox{90}{\textbf{dSprites/ori}}   & \rotatebox{90}{\textbf{SmallNORB/azi}}  & \rotatebox{90}{\textbf{SmallNORB/ele}} & \rotatebox{90}{\textbf{Mean}} & \rotatebox{90}{\textbf{\# Params. (M)}} & \rotatebox{90}{\textbf{PPT}}\\
			\midrule
                \multicolumn{23}{c}{\textit{\textbf{Conventional Finetuning}}} \tabularnewline \midrule
		\rowcolor{cyan!8}Full fine-tuning \cite{jia2022visual}  & 68.9 & 87.7 & 64.3 & 97.2  & 86.9 &87.4 & 38.8 & 79.7 &95.7 &84.2 & 73.9 & 56.3 & 58.6  & 41.7 & 65.5 & 57.5 & 46.7 & 25.7 & 29.1 &  65.57  & 85.8M & - \\ 
		Linear probing \cite{jia2022visual}  & 63.4  & 85.0 & 63.2          & 97.0 & 86.3 & 36.6  & 51.0 & 78.5 & 87.5 & 68.6  &74.0     & 34.3 & 30.6 & 33.2 & 55.4 & 12.5 & 20.0 & 9.6 & 19.2 & 52.94 &\textcolor{red}{0M} & 0.53\\  \midrule
                \multicolumn{23}{c}{\textit{\textbf{PEFT Algorithms}}} \tabularnewline \midrule

            \rowcolor{cyan!8}Adapter~\cite{houlsby2019parameter}   & 69.2 & 90.1 & 68.0 & 98.8 & 89.9 & 82.8 & 54.3 & 84.0 & 94.9 & 81.9 & 75.5 & 80.9 & 65.3 & 48.6 & 78.3 & 74.8 & 48.5 & 29.9 & 41.6 & 71.44 & 0.16M & 0.71\\
  
                
		VPT-Shallow \cite{jia2022visual} &77.7 & 86.9 & 62.6& 97.5& 87.3& 74.5& 51.2& 78.2& 92.0& 75.6& 72.9& 50.5& 58.6& 40.5& 67.1 & 68.7& 36.1& 20.2& 34.1 &64.85 & 0.08M & 0.65\\ 
  
		\rowcolor{cyan!8}VPT-Deep \cite{jia2022visual} &78.8 &90.8 &65.8 & 98.0 &88.3 & 78.1& 49.6 &81.8 &96.1 & 83.4& 68.4 &68.5 &60.0 &46.5 &72.8 &73.6 &47.9 &32.9 &37.8 &69.43 & 0.56M & 0.68\\ 
            
            BitFit \cite{zaken2021bitfit}  &  72.8 & 87.0 & 59.2 & 97.5 & 85.3 & 59.9 &51.4 & 78.7& 91.6& 72.9& 69.8& 61.5 & 55.6&32.4 & 55.9& 66.6& 40.0& 15.7& 25.1 &62.05 & 0.10M & 0.61\\

            \rowcolor{cyan!8}LoRA~\cite{hu2021lora}   & 67.1 & 91.4 & 69.4 & 98.8 & 90.4 & 85.3 & 54.0 & 84.9 & 95.3 & 84.4 & 73.6 & 82.9 & \textcolor{red}{69.2} & 49.8 & 78.5 & 75.7 & 47.1 & 31.0 & 44.0 &72.25 & 0.29M &0.71\\

            AdaptFormer~\cite{chen2022adaptformer}   & 70.8 & 91.2 & 70.5 & 99.1 & 90.9 & 86.6 & 54.8 & 83.0 & 95.8 & 84.4 &\textcolor{citecolor}{76.3} & 81.9 & 64.3 & 49.3 & 80.3 & 76.3 & 45.7  & 31.7 & 41.1 &72.32 & 0.16M & 0.72\\

            \rowcolor{cyan!8}SSF~\cite{lian2022scaling}  &69.0 &92.6 &75.1 &\textcolor{citecolor}{99.4} &\textcolor{red}{91.8} & 90.2 &52.9 &\textcolor{citecolor}{87.4} &95.9 &\textcolor{citecolor}{87.4} &75.5 &75.9 &62.3 &\textcolor{citecolor}{53.3} &80.6 & 77.3 &54.9 &29.5 &37.9 &73.10  &0.21M & 0.72\\

            NOAH~\cite{liu2022neural}   & 69.6 &92.7 & 70.2 & 99.1 & 90.4 & 86.1 & 53.7 & 84.4 & 95.4 & 83.9 & 75.8 & 82.8 & \textcolor{citecolor}{68.9} & 49.9 & 81.7 & 81.8 & 48.3 & 32.8 & 44.2 &73.25 & 0.43M & 0.72\\

            \rowcolor{cyan!8}SCT\cite{zhao2023sct} &75.3 &91.6 &72.2 &99.2 &91.1 &\textcolor{citecolor}{91.2} &55.0 &85.0 &96.1 &86.3 &76.2 &81.5 &65.1 &51.7 &80.2 &75.4 &46.2 &33.2 &45.7 &73.59 &0.11M & 0.73 \\

            FacT~\cite{jie2023fact}  & 70.6 & 90.6 & 70.8 & 99.1 & 90.7 & 88.6 & 54.1 & 84.8 &96.2 & 84.5 & 75.7 & 82.6 & 68.2 & 49.8 & 80.7 & 80.8 & 47.4 & 33.2 & 43.0 &73.23 & 0.07M & 0.73\\

            \rowcolor{cyan!8}RepAdapter~\cite{luo2023towards} & 72.4 & 91.6 & 71.0 & 99.2 & 91.4 &90.7 & 55.1 & 85.3 & 95.9 & 84.6 & 75.9 & 82.3 & 68.0 & 50.4 & 79.9 & 80.4 & 49.2 &\textcolor{red}{38.6} & 41.0 &73.84 & 0.22M & 0.72\\

            Hydra~\cite{kim2023hydra}  & 72.7 & 91.3 & 72.0 & 99.2 & 91.4 &90.7 &\textcolor{citecolor}{55.5} & 85.8 & 96.0 & 86.1 & 75.9 &\textcolor{citecolor}{83.2} & 68.2 & 50.9 &82.3 & 80.3 & 50.8 & 34.5 & 43.1 &74.21 &0.28M &0.73\\

             \rowcolor{cyan!8}LST~\cite{sung2022lst} &59.5 &91.5 &69.0 &99.2 &89.9 &79.5 &54.6 &86.9 &95.9 &85.3 &74.1 &81.8 &61.8 &52.2 &81.0 &71.7 &49.5 &33.7 &45.2 &71.70 &2.38M & 0.65\\

             DTL~\cite{fu2024dtl} &69.6 &\textcolor{red}{94.8} &71.3 &99.3 &91.3 &83.3 &\textcolor{red}{56.2} &87.1 &96.2 &86.1 &75.0 &82.8 &64.2 &48.8 &81.9 &\textcolor{red}{93.9} &53.9 &34.2 &47.1 &74.58 & \textcolor{citecolor}{0.04M} &\textcolor{red}{0.75}\\

              \rowcolor{cyan!8}HST~\cite{lin2023hierarchical}&76.7 &94.1 &74.8 &\textcolor{red}{99.6} &91.1 &\textcolor{citecolor}{91.2} &52.3 &87.1 &\textcolor{citecolor}{96.3} &\textcolor{red}{88.6} &\textcolor{red}{76.5} &\textcolor{red}{85.4} &63.7 &52.9 &81.7 &87.2 &\textcolor{red}{56.8} &\textcolor{citecolor}{35.8} &\textcolor{red}{52.1} &\textcolor{red}{75.99} & 0.78M & \textcolor{citecolor}{0.74}\\

               GPS~\cite{zhang2024gradient}&\textcolor{citecolor}{81.1} &\textcolor{citecolor}{94.2} &\textcolor{citecolor}{75.8} &\textcolor{citecolor}{99.4} &\textcolor{citecolor}{91.7} &\textcolor{red}{91.6} &52.4 &\textcolor{red}{87.9} &96.2 &86.5 &\textcolor{red}{76.5} &79.9 &62.6 &\textcolor{red}{55.0} &82.4 &84.0 &\textcolor{citecolor}{55.4} &29.7 &46.1 &\textcolor{citecolor}{75.18} &0.22M & \textcolor{citecolor}{0.74}\\
               
                \rowcolor{cyan!8}LAST~\cite{tang2024low} &66.7 &93.4 &\textcolor{red}{76.1} &\textcolor{red}{99.6} &89.8 & 86.1& 54.3 & 86.2 &\textcolor{citecolor}{96.3} & 86.8 &75.4 &81.9 &65.9 &49.4 &\textcolor{red}{82.6} &\textcolor{citecolor}{87.9} &46.7 &32.3 &\textcolor{citecolor}{51.5} &74.15 &0.66M & 0.72
               \\

                SNF~\cite{wang2023adapting} &\textcolor{red}{84.0} & 94.0 &72.7 &99.3 &91.3 &90.3 &54.9 &87.2 &\textcolor{red}{97.3} &85.5 &74.5 &82.3 &63.8 &49.8 &\textcolor{citecolor}{82.5} &75.8 &49.2 &31.4 &42.1 &74.10 &0.25M & 0.73\\
            \bottomrule
		\end{tabular}
	}
	\label{table: vtab}
    \vspace{-0.15in}
\end{table*}
\vspace{0.1cm} \noindent \textbf{Core Layer.} In this layer, we implement the essential functions commonly used to train PEFT algorithms. Additionally, this layer includes the code for datasets, data loaders, and pre-trained models that are utilized in the V-PEFT Bench.

\vspace{0.1cm} \noindent \textbf{Algorithm Layer.} In the algorithm layer, we implement the base class for PEFT algorithms, which includes initializing the datasets, data loaders, and models from the core layer. Moreover, we implement the loss functions and algorithm-specific configurations used in PEFT algorithms. Based on these implementations, we support 25 PEFT algorithms in the V-PEFT Bench. More algorithms are expected to be added through the continued extension of V-PEFT Bench.

\vspace{0.1cm} \noindent \textbf{Extension Layer.} The extension layer is dedicated to advancing the core PEFT algorithms for visual analysis. In this layer, we  implement attention map and feature distribution visualization, enabling researchers to directly observe and compare the performance of different PEFT algorithms.

\vspace{0.1cm} \noindent \textbf{API Layer.} We encapsulate the core functions and algorithms within the API layer, creating a user-friendly interface to apply PEFT algorithms to new applications. In addition, we provide configuration files for all supported algorithms, complete with detailed parameter settings, which enable the reproduction of results.


\vspace{-0.2in}
\subsection{PEFT Algorithms Implemented in the V-PEFT Bench}
We implement 2 conventional and 25 PEFT algorithms in the codebase for the V-PEFT Bench, including full fine-tuning, Frozen, Adapter~\cite{houlsby2019parameter}, AdaptFormer~\cite{chen2022adaptformer}, SCT~\cite{zhao2023sct}, BitFit~\cite{zaken2021bitfit}, U-Tuning~\cite{jiang2023rethinking}, VPT-shallow~\cite{jia2022visual}, VPT-Deep~\cite{jia2022visual}, Prefix Tuning~\cite{yu2022towards}, SSF~\cite{lian2022scaling}, LoRA~\cite{hu2021lora}, NOAH~\cite{liu2022neural}, FacT~\cite{jie2023fact}, RepAdapter~\cite{luo2023towards}, Hydra~\cite{kim2023hydra}, LST~\cite{sung2022lst}, DTL~\cite{fu2024dtl}, HST~\cite{lin2023hierarchical}, GPS~\cite{zhang2024gradient}, LAST~\cite{tang2024low}, SNF~\cite{wang2023adapting}, BAPAT~\cite{yu2022towards}, LN TUNE~\cite{basu2024strong}, LoRand~\cite{yin20231}, E$^3$VA~\cite{yin2023parameter}, and Mona~\cite{yin2023adapter}.  
The algorithms are chosen based on the following considerations: 1) As detailed in Section~\ref{sec:taxonomy}, existing PEFT algorithms are categorized into 7 basic categories. For each category, we implement 2 to 5 representative algorithms; 2) The algorithm is widely adopted in the visual PEFT domain; 3) The algorithm corresponds with the comprehensive timeline of visual PEFT development. 
\vspace{-0.2in}
\subsection{Unified Evaluation Metrics} 
In PEFT, the evaluation of algorithms typically focuses on two aspects: the number of trainable parameters and task performance. Algorithms that achieve superior performance with fewer trainable parameters generally attract more attention. However, currently there are no standardized metrics to comprehensively evaluate PEFT algorithms. To address this gap, we propose a unified \textbf{Performance-Parameter Trade-off} (\textbf{PPT}) metric, which shows how each increase in trainable parameters improves the performance of downstream tasks. Specifically, the PPT metric for a PEFT algorithm $M$ considers its performance $M_t$ on a downstream task $t$, its trainable parameters $P_M$, and a normalization constant $C$. The formula for $PPT_M$ is defined as:
\vspace{-0.1in}
\begin{equation}
    PPT_M = M_t \times \exp(-\log_{10}(\frac{P_M}{C}+1)).
\end{equation}
Special attention is required for the parameters because they range from $0M$ to $10M$ ($1M=10^6$) in the PEFT algorithms. Therefore, the normalization constant $C$ is set to $10^7$, aligning with the typical parameter scales of existing PEFT algorithms. This ensures that the ratio $\frac{P_M}{C}$ remains within the range [0, 1). Additionally, to prevent the logarithmic term from reaching negative infinity, we introduce an additive constant of 1.
\vspace{-0.1in}
\subsection{Image Recognition Results}


Table~\ref{table: vtab} presents the benchmark results for 18 PEFT algorithms on VTAB \cite{jia2022visual}. 
These evaluation results show that : \textbf{(1)} Almost all PEFT algorithms outperform full fine-tuning, demonstrating that fully fine-tuning the pre-trained ViT on limited data risks overfitting and catastrophic forgetting. In contrast, fine-tuning only a few parameters helps maintain the generalizability of the pre-trained models when adapting to downstream tasks. \textbf{(2)} DTL~\cite{fu2024dtl} achieves the best PPT by leveraging low-rank linear mappings and feature reuse to reduce tunable parameters while enhancing performance. \textbf{(3)} Most PEFT algorithms perform well on the \textbf{\textit{Natural}} and \textbf{\textit{Specialized}} groups because their classification objectives align with the training goals of the pre-trained dataset, ImageNet-21K \cite{ridnik2021imagenet}. 
However, the \textbf{\textit{Structured}} group tasks, such as object counting and depth prediction, differ significantly from 
training objectives on ImageNet, resulting in a substantial domain gap. PEFT algorithms with less extensive parameter tuning, such as BitFit \cite{zaken2021bitfit} and VPT-Shallow \cite{jia2022visual}, fail to bridge this gap, leading to suboptimal performance.


\setlength{\tabcolsep}{9pt} 
\renewcommand{\arraystretch}{1} 
\begin{table*}[ht]
 \vspace{-0.1in}
  \caption{\textbf{Benchmark Results on SSv2 and HMDB51.} We evaluate 2 conventional fine-tuning algorithms and 5 PEFT algorithms with ViT-B from VideoMAE and Video Swin transformer. We highlight the \textcolor{red}{best} and the \textcolor{citecolor}{second} results.}
  \label{tab:video}
\begin{center}
\vspace{-0.15in}
    \resizebox{0.9\linewidth}{!}{
     {\renewcommand{\arraystretch}{1}
  \begin{tabular}{cccccccc}
     \toprule
     \multirow{2}[3]{*}{\textbf{Method}} & \multirow{2}[3]{*}{\textbf{Model}}  & \multirow{2}[3]{*}{\textbf{Pre-training}} &\multirow{2}[3]{*}{\textbf{\# Params.}} & \multicolumn{2}{c}{\textbf{SSv2}}
     &\multicolumn{2}{c}{\textbf{HMDB51}}\\
     
    \cmidrule(lr){5-6}
    \cmidrule(lr){7-8}
    & & & &Top1 & PPT &Top1 & PPT \\
    \midrule
    \multicolumn{8}{c}{\textit{\textbf{Vision transformer (from VideoMAE)}}}\\ \midrule
    
    \rowcolor{cyan!8}Full fine-tuning &ViT-B & Kinetics 400  & 85.97 M  &  53.97 \% & - & 46.41 \%    & -  \\

    Frozen &ViT-B & Kinetics 400 &\textcolor{red}{0 M}  & 29.23 \% & 0.29 & 49.84 \%  & \textcolor{citecolor}{0.50} \\

    \rowcolor{cyan!8}AdaptFormer \cite{chen2022adaptformer} &ViT-B & Kinetics 400 & \textcolor{citecolor}{1.19 M}  &\textcolor{red}{59.02 \%}  &\textcolor{red}{0.56} &  \textcolor{citecolor}{55.69 \%}  &\textcolor{red}{0.53}  \\

    BAPAT~\cite{yu2022towards} &ViT-B & Kinetics 400    & 2.06 M  & \textcolor{citecolor}{57.78 \%}  & \textcolor{citecolor}{0.53} &\textcolor{red}{57.18 \%}   &\textcolor{red}{0.53}  \\ 
    
    \midrule
    \multicolumn{8}{c}{\textit{\textbf{Video Swin transformer}}}\\ \midrule

    \rowcolor{cyan!8}Full fine-tuning &Video Swin-B & Kinetics 400  & 87.64 M  & \textcolor{citecolor}{50.99 \%} & - &  68.07 \%   &  - \\
    
    Frozen &Video Swin-B & Kinetics 400 &\textcolor{red}{0 M}  &  24.13 \% & 0.24 & \textcolor{citecolor}{71.28 \%}   &\textcolor{red}{0.71} \\
    
    \rowcolor{cyan!8}LoRA \cite{hu2021lora} &Video Swin-B & Kinetics 400 & \textcolor{citecolor}{0.75 M}  &38.34 \%  & 0.37 &  62.12 \%    & 0.60 \\ 
      
    BitFit \cite{zaken2021bitfit} &Video Swin-B & Kinetics 400 & 1.09 M  &  45.94 \% &\textcolor{red}{0.44}  &  68.26 \%   &  \textcolor{citecolor}{0.65} \\

    \rowcolor{cyan!8}AdaptFormer \cite{chen2022adaptformer} &Video Swin-B & Kinetics 400 & 1.56 M  &  40.80 \% & 0.38 &  68.66 \%    &  0.64  \\
    
    Prefix-tuning \cite{li2021prefix} &Video Swin-B & Kinetics 400 & 6.37 M  & 39.46 \% &  0.32 & 56.13 \%   & 0.45 \\
    
    \rowcolor{cyan!8}BAPAT~\cite{yu2022towards} &Video Swin-B & Kinetics 400    & 6.18 M  &\textcolor{red}{53.36 \%}  & \textcolor{citecolor}{0.43} &\textcolor{red}{71.93 \%}    & 0.58   \\
     \bottomrule
  \end{tabular}
  }
  }
  \end{center}
\end{table*}

\setlength{\tabcolsep}{3pt} 
\renewcommand{\arraystretch}{1} 
\begin{table*}[t]
	\centering
    \vspace{-0.1in}
        \caption{\textbf{Benchmark Results on MS COCO, PASCAL VOC, and ADE20K.} We evaluate 9 PEFT algorithms using Swin-B  pre-trained on ImageNet-22K for MS COCO and Swin-L pre-trained on ImageNet-22K for PASCAL VOC and ADE20K. We highlight the \textcolor{red}{best} and the \textcolor{citecolor}{second} results.}
        \label{tab:coco}
        \vspace{-0.1in}
	\resizebox{1.0\linewidth}{!}{
			\begin{tabular}{@{}crrcccccccccc@{}}
				\toprule
                    & \multicolumn{6}{c}{\textbf{\textit{Swin-B}}} &\multicolumn{6}{c}{\textbf{\textit{Swin-L}}}\\
                    \cmidrule(lr){2-7}
                     \cmidrule(lr){8-13}
			
                \multirow{2}{*}{\textbf{\begin{tabular}[c]{@{}l@{}}~~Method\end{tabular}}} & \multicolumn{1}{c}{\multirow{2}{*}{\textbf{\begin{tabular}[c]{@{}c@{}}\# Params. \end{tabular}}}} 
                
                & \multicolumn{1}{c}{\multirow{2}{*}{\textbf{\begin{tabular}[c]{@{}c@{}} Memory\end{tabular}}}} &  \multicolumn{4}{c}{\textbf{\begin{tabular}[c]{@{}c@{}}COCO\\ (Cascade Mask R-CNN)\end{tabular}}} 

                & \multicolumn{1}{c}{\multirow{2}{*}{\textbf{\begin{tabular}[c]{@{}c@{}}\# Params. \end{tabular}}}}  & 
				\multicolumn{1}{c}{\multirow{2}{*}{\textbf{\begin{tabular}[c]{@{}c@{}}Memory \end{tabular}}}} & \multicolumn{2}{c}{\textbf{\begin{tabular}[c]{@{}c@{}}Pascal VOC\\ (RetinaNet)\end{tabular}}} & \multicolumn{2}{c}{\textbf{\begin{tabular}[c]{@{}c@{}}ADE20K\\ (UPerNet)\end{tabular}}} \\
                
                \cline{4-7} 
                \cline{10-13} 
			& & & \multicolumn{1}{c}{$\bm{\mathrm{AP_{Box}}}$} &PPT & \multicolumn{1}{c}{$\bm{\mathrm{AP_{Mask}}}$} &PPT  & & & \multicolumn{1}{c}{$\bm{\mathrm{AP_{Box}}}$} & PPT & \multicolumn{1}{c}{$\bm{\mathrm{mIoU}}$} & PPT\\ \midrule
				
			\multicolumn{13}{c}{\textit{\textbf{Conventional Finetuning}}}\\ \midrule
				
				\rowcolor{cyan!8}Full fine-tuning & \multicolumn{1}{r}{86.75 M} & 17061 MB& \multicolumn{1}{c}{\textcolor{citecolor}{51.9 \%}} & - & \multicolumn{1}{c}{\textcolor{citecolor}{45.0 \%}} & - & \multicolumn{1}{r}{198.58 M} & 15679 MB & \multicolumn{1}{c}{83.5 \%} & - & \multicolumn{1}{c}{\textcolor{red}{52.1 \%}} & \multicolumn{1}{c}{-} \\
    
			Frozen & \multicolumn{1}{r}{\textcolor{red}{0.00 M}} & \textcolor{red}{7137 MB} & \multicolumn{1}{c}{43.5 \%} & \multicolumn{1}{r}{0.44} & \multicolumn{1}{c}{38.6 \%} & \multicolumn{1}{r}{0.39} & \multicolumn{1}{r}{\textcolor{red}{0.00 M}} & \textcolor{citecolor}{3967 MB} & \multicolumn{1}{c}{83.6 \%} & \multicolumn{1}{r}{0.84} & \multicolumn{1}{c}{46.8 \%} &  \multicolumn{1}{r}{\textcolor{citecolor}{0.47}}\\
    
                \midrule
                \multicolumn{13}{c}{\textit{\textbf{PEFT Algorithms}}}\\ \midrule
    
			\rowcolor{cyan!8}Bitfit~\cite{zaken2021bitfit} & \multicolumn{1}{r}{0.20 M} &13657 MB & \multicolumn{1}{c}{47.9 \%} & \multicolumn{1}{r}{0.47} & \multicolumn{1}{c}{41.9 \%} & \multicolumn{1}{r}{\textcolor{red}{0.42}} & \multicolumn{1}{r}{0.30 M} &10861 MB & \multicolumn{1}{c}{85.7 \%} & \multicolumn{1}{r}{\textcolor{citecolor}{0.85}} &\multicolumn{1}{c}{48.3 \%}&\multicolumn{1}{r}{\textcolor{red}{0.48}}\\
    
			LN TUNE~\cite{basu2024strong} & \multicolumn{1}{r}{\textcolor{citecolor}{0.06 M}} &12831 MB & \multicolumn{1}{c}{48.0 \%} & \multicolumn{1}{r}{\textcolor{citecolor}{0.48}} & \multicolumn{1}{c}{41.4 \%} & \multicolumn{1}{r}{\textcolor{citecolor}{0.41}} & \multicolumn{1}{r}{\textcolor{citecolor}{0.09 M}} &10123 MB & \multicolumn{1}{c}{85.8 \%} & \multicolumn{1}{r}{\textcolor{red}{0.86}} &\multicolumn{1}{c}{47.9 \%}&\multicolumn{1}{r}{\textcolor{red}{0.48}}\\
    
   
			Adapter~\cite{houlsby2019parameter} & \multicolumn{1}{r}{3.11 M} & 12557 MB & \multicolumn{1}{c}{50.9 \%} & \multicolumn{1}{r}{0.45} & \multicolumn{1}{c}{43.8 \%} & \multicolumn{1}{r}{0.39} & \multicolumn{1}{r}{4.66 M} &10793 MB & \multicolumn{1}{c}{87.1 \%} & \multicolumn{1}{r}{0.74} &\multicolumn{1}{c}{50.7 \%}&\multicolumn{1}{r}{0.43}\\
    
			\rowcolor{cyan!8}LoRA~\cite{hu2021lora} & \multicolumn{1}{r}{3.03 M} &11975 MB & \multicolumn{1}{c}{51.2 \%} & \multicolumn{1}{r}{0.46} & \multicolumn{1}{c}{44.3 \%} & \multicolumn{1}{r}{0.40} & \multicolumn{1}{r}{4.57 M} &10127 MB & \multicolumn{1}{c}{\textcolor{red}{87.5 \%}} & \multicolumn{1}{r}{0.74} &\multicolumn{1}{c}{50.3 \%} &\multicolumn{1}{r}{0.43}\\

			AdaptFormer~\cite{chen2022adaptformer} & \multicolumn{1}{r}{3.11 M} &13186 MB & \multicolumn{1}{c}{51.4 \%} & \multicolumn{1}{r}{0.46} & \multicolumn{1}{c}{44.5 \%} & \multicolumn{1}{r}{0.40} & \multicolumn{1}{r}{4.66 M} &11036 MB & \multicolumn{1}{c}{\textcolor{citecolor}{87.3 \%}} & \multicolumn{1}{r}{0.74} &\multicolumn{1}{c}{50.8 \%}&\multicolumn{1}{r}{0.43}\\

            \rowcolor{cyan!8}LoRand~\cite{yin20231} & 1.20 M &13598 MB &\multicolumn{1}{c}{51.0 \%} &\multicolumn{1}{r}{\textcolor{red}{0.49}} & \multicolumn{1}{c}{43.9 \%} &\multicolumn{1}{c}{\textcolor{red}{0.42}} & \multicolumn{1}{r}{1.31 M} &11572 MB & \multicolumn{1}{c}{86.8 \%} &\multicolumn{1}{r}{0.82} & 50.7 \% &\multicolumn{1}{r}{\textcolor{red}{0.48}} \\
			
			E$^3$VA~\cite{yin2023parameter} & \multicolumn{1}{r}{1.20 M} & 7639 MB & \multicolumn{1}{c}{50.5 \%} & \multicolumn{1}{r}{\textcolor{citecolor}{0.48}} & \multicolumn{1}{c}{43.8 \%} &  \multicolumn{1}{r}{\textcolor{red}{0.42}} & \multicolumn{1}{r}{1.79 M} &4819 MB & \multicolumn{1}{c}{86.5 \%} & \multicolumn{1}{r}{0.81} &\multicolumn{1}{c}{49.6 \%}&\multicolumn{1}{r}{0.46}\\

            \rowcolor{cyan!8}Mona~\cite{yin2023adapter} & 4.16 M &13996 MB & \multicolumn{1}{c}{\textcolor{red}{53.4 \%}} &\multicolumn{1}{r}{0.46} &\multicolumn{1}{c}{\textcolor{red}{46.0 \%}} &\multicolumn{1}{r}{0.40} &\multicolumn{1}{r}{5.08 M} &11958 MB &\multicolumn{1}{c}{\textcolor{citecolor}{87.3 \%}} &\multicolumn{1}{r}{0.73} &\multicolumn{1}{c}{\textcolor{citecolor}{51.3 \%}} &\multicolumn{1}{r}{0.43}\\
			\bottomrule
	\end{tabular}
        }
	\label{tab:voc and ade20k}
    \vspace{-0.15in}
\end{table*}

\begin{figure*}[t]
    \centering
    \vspace{-0.12in}
    \includegraphics[width=1\linewidth]{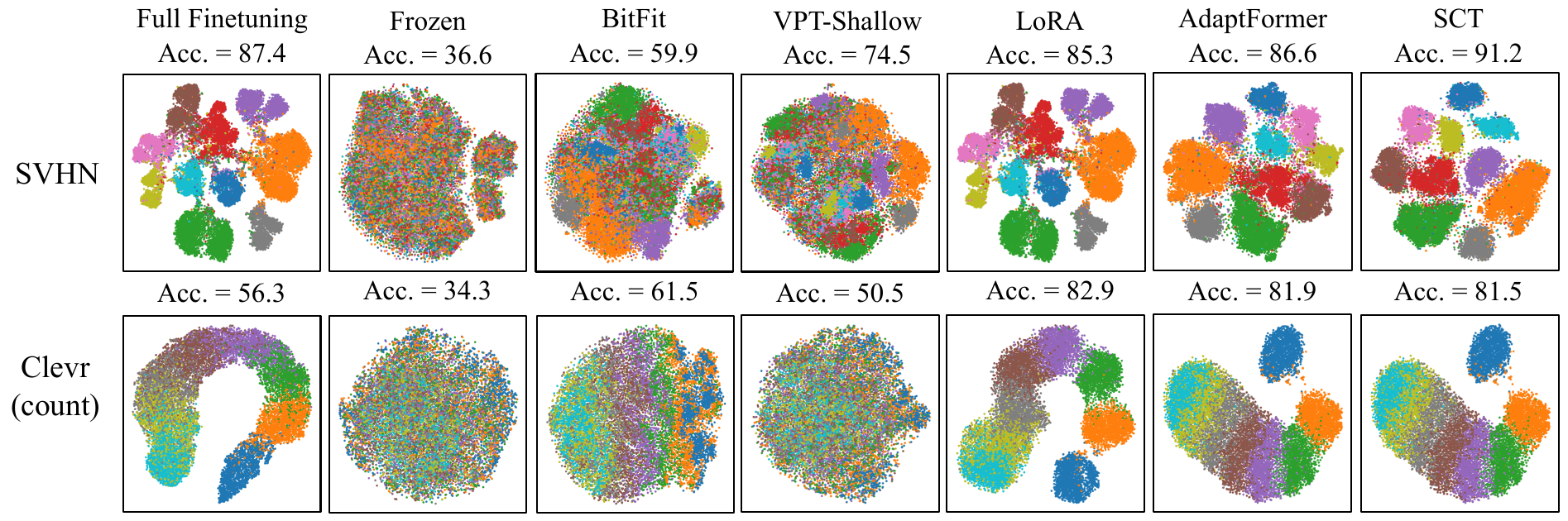}
    \vspace{-0.2in}
    \caption{
    \textbf{Visualization of Feature Distribution on SVHN and Clevr/count.}
    The more distinguishable the feature distributions, the better the method’s performance.}
   \vspace{-0.15in}
    \label{fig:tsne}
\end{figure*}



\begin{figure*}[t]
    \centering
    \includegraphics[width=1\linewidth]{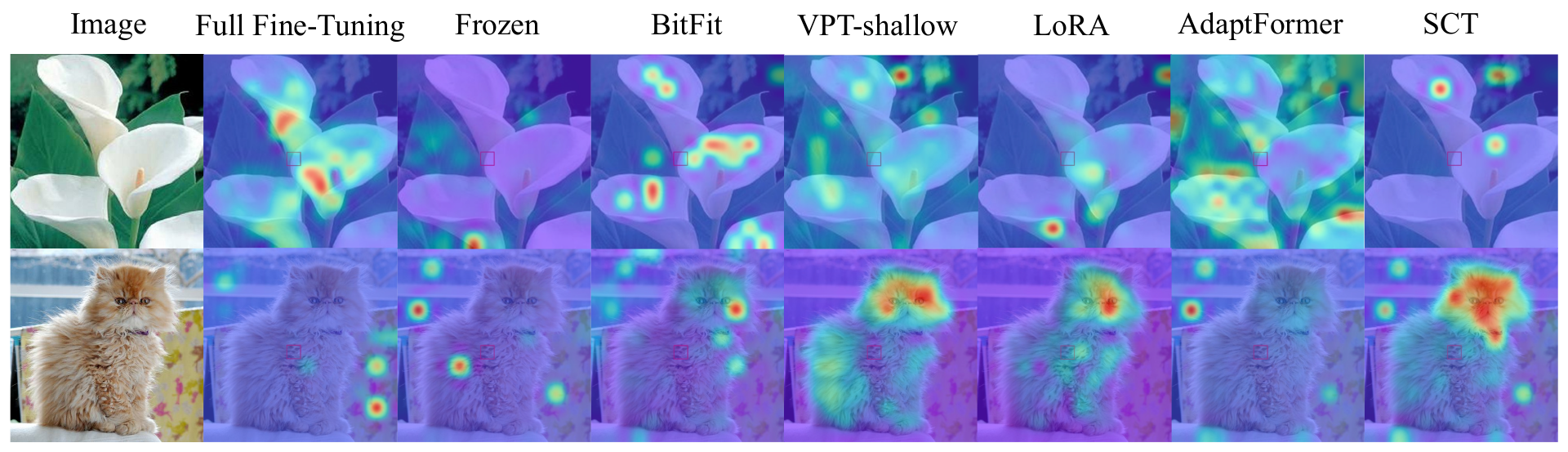}
    \vspace{-0.2in}
    \caption{
    \textbf{Visualization of Attention Maps.} From left to right, each column shows the RGB image and various PEFT methods.
    }
    \label{fig:attention}
    \vspace{-0.15in}
\end{figure*}

\vspace{-0.17in}
\subsection{Video Action Recognition Results} 
Table~\ref{tab:video} displays evaluation results for 5 PEFT algorithms using ViT-B from VideoMAE~\cite{tong2022videomae} and Video Swin transformer~\cite{liu2022video} on the SSv2 \cite{goyal2017something} and HMDB51 \cite{kuehne2011hmdb} datasets. 
The findings are as follows: \textbf{(1)} On SSv2, which has sufficient data, the ViT-B from VideoMAE outperforms others, illustrating the robustness of features learned through self-supervised learning and the improved generalization of the pre-trained model. Conversely, on HMDB51, which has limited data and fewer categories, the supervised pre-trained Video Swin transformer shows superior performance, indicating better adaptability and generalization in smaller datasets. \textbf{(2)} On SSv2, only a few PEFT algorithms outperform full fine-tuning, suggesting that with sufficient data, full fine-tuning is less likely to overfit. Conversely, on HMDB51, most PEFT algorithms outperform full fine-tuning, indicating that full fine-tuning may lead to overfitting when data are scarce, whereas PEFT algorithms offer a more effective solution. \textbf{(3)} BAPAT \cite{yu2022towards} achieves outstanding performance by integrating the strengths of Adapter \cite{houlsby2019parameter}, Prefix \cite{li2021prefix}, and Prompt \cite{jia2022visual}.

\vspace{-0.15in}
\subsection{Dense Prediction Results}
\subsubsection{Benchmark Results on MS-COCO} Table~\ref{tab:coco} shows the results on MS-COCO \cite{lin2014microsoft} using 9 PEFT algorithms with pre-trained Swin-B. Our analysis reveals that: \textbf{(1)} Full fine-tuning generally outperforms most PEFT algorithms. This is because MS-COCO is a substantial dataset with sufficient data, reducing the likelihood of overfitting when fully fine-tuning. However, most PEFT algorithms show competitive performance, demonstrating their parameter efficiency. \textbf{(2)} Mona \cite{yin2023adapter} stands out as the only PEFT algorithm to surpass full fine-tuning, showcasing the effectiveness of its multi-cognitive visual filters.

\vspace{-0.1in}
\subsubsection{Benchmark Results on PASCAL VOC and ADE20K} Table~\ref{tab:voc and ade20k} presents the results on PASCAL VOC \cite{everingham2015pascal} and ADE20K \cite{zhou2017scene} using 9 PEFT algorithms. We can observe that: \textbf{(1)} On PASCAL VOC, which features fewer data and object categories, all PEFT algorithms surpass full fine-tuning. This is because adjusting a small number of parameters in the pre-trained model helps prevent overfitting and catastrophic forgetting, thereby preserving the model's generalization ability. 
On the other hand, on ADE20K, which has more data and object categories, full fine-tuning outperforms all PEFT algorithms. With more available data, fully fine-tuning the pre-trained model allows for better adaptation to the downstream task. However, PEFT algorithms still achieve competitive results, demonstrating their parameter efficiency. \textbf{(2)} LN TUNE \cite{basu2024strong} achieves the highest performance on both PASCAL VOC and ADE20K, indicating that fine-tuning only the LayerNorm parameters is effective and efficient.


\vspace{-0.15in}
\subsection{Visualization}
In addition to the quantitative analysis of PEFT algorithms, V-PEFT Bench provides t-SNE and attention map visualizations to facilitate a deeper understanding of PEFT algorithms.
\vspace{-0.1in}
\subsubsection{t-SNE for Feature Distribution Visualization} V-PEFT Bench offers t-SNE visualizations that intuitively represent the distribution of features for downstream tasks. These visualizations allow us to assess the effectiveness of the PEFT algorithms. Figure~\ref{fig:tsne} shows the t-SNE visualizations for two specific tasks in VTAB-1k, SVHN, and Clevr/count. The visualizations demonstrate that the feature distribution of the data is closely linked to performance, with higher performance showing more distinct decision boundaries.

\vspace{-0.1in}
\subsubsection{Attention Map}
Attention map visualization is a crucial tool for analyzing PEFT algorithms. In the V-PEFT Bench, we have included an attention map visualization module. We randomly select several examples of attention map visualizations, as illustrated in Figure~\ref{fig:attention}. We can find that the SCT method focuses more intently on the cat, while full fine-tuning directs more concentrated attention toward the flower. In the V-PEFT Bench, we offer a convenient interface that allows researchers to easily utilize the attention map visualization tool.


\vspace{-0.15in}
\subsection{Future Development of V-PEFT Bench}
The V-PEFT Bench is a major undertaking and resource-consuming project. However, we envision the V-PEFT Bench as a long-term, evolving initiative and are committed to its continuous development. Our future roadmap includes broadening its scope to encompass more CV tasks and incorporating a diverse range of pre-trained vision models. Currently, the V-PEFT Bench focuses primarily on conventional CV tasks. However, there are additional CV tasks that merit the attention of the PEFT community. Currently, V-PEFT Bench does not cover tasks such as text-to-image generation, point cloud analysis, or robotic manipulation, which are critical areas in modern CV applications. Furthermore, the benchmark currently supports a limited selection of pre-trained models. Expanding the repository to include self-supervised models, multimodal pre-trained models, and diffusion models is a key priority. We aim to address these limitations by continuously updating the benchmark to incorporate these enhancements.

\section{Future Research Challenges}
\label{sec:direction}
\subsection{Explainability of Visual PEFT Methods}
Despite significant advancements, the underlying reasons for the effectiveness of visual PEFT methods remain unclear, especially in terms of the interpretability of visual prompts. 
In the NLP domain, we can explain the prompt with a better description, which is more intuitive.
While in the vision domain, the main challenge is that visual prompts are learned as unordered token-based prompts, which are difficult to translate into an understandable format. Other tuning techniques such as adapter and prefixes also face challenges in interpretability. These methods strive to reduce the number of parameters required for adapting large models to specific tasks. Therefore, improving the interpretability of PEFT is a crucial area for future research.

\subsection{PEFT for Visual Generative Models}
Within the CV domain, most PEFT methods are designed primarily for discriminative tasks, such as image classification and recognition of video actions. Although several PEFT approaches have been proposed for pre-trained diffusion generative models, these methods remain relatively underdeveloped, leaving significant potential for further investigation. Beyond diffusion models, autoregressive generative models (AR) have also recently gained traction in the field of visual generation. By discretizing images and employing AR models to learn image sequences, this paradigm offers a novel approach to visual generation. However, to date, there has been little to no research on the application of PEFT to AR models. 
As a result, advancing PEFT techniques for both diffusion and AR generative models presents a highly promising direction.

\vspace{-0.2in}
\subsection{Simplify Hyperparameter Selection}
Through our experiment in constructing the benchmark, we observe that the performance of existing PEFT methods is heavily influenced by hyperparameter configurations, which can be categorized into two primary aspects: 1) Method-specific hyperparameters: The efficacy of PEFT methods is often highly sensitive to their intrinsic hyperparameters, such as the bottleneck dimension of the adapter, the rank of LoRA, and the arrangement of various additive PEFT layers. 2) Training hyperparameters: Certain training hyperparameters (e.g., learning rate, weight decay, etc.) exhibit extreme sensitivity in some methods, where even minor adjustments can result in significant performance variations. Manual tuning of these hyperparameters demands considerable effort and resources. Consequently, future research could focus on devising approaches that reduce the dependence on manual hyperparameter optimization or enable the automatic discovery of optimal configurations.

\vspace{-0.15in}
\subsection{Differential Privacy for PEFT methods}
Different downstream tasks frequently involve varying degrees of sensitive and personal data, underscoring the critical importance of ensuring privacy during model fine-tuning, especially when employing PEFT methods. The integration of large-scale model fine-tuning with differential privacy presents a compelling avenue for future research. However, current differential privacy techniques, such as DP-AdamW~\cite{lilarge}, often suffer from constrained performance and considerable computational overhead. Consequently, future research should prioritize the development of efficient, scalable, and privacy-preserving approaches specifically tailored to the unique requirements of PEFT methods.

\vspace{-0.15in}
\section{Conclusion}
In this paper, we present a comprehensive review of the visual PEFT domain, providing an in-depth analysis of existing methods, datasets, and applications. We conclude with a thorough comparison of these methods and highlight several key research challenges within the field. Furthermore, we introduce a unified and challenging PEFT benchmark for CV tasks, designed to enable fair and consistent evaluations. The objective of this survey and benchmark is to serve as a valuable resource for researchers in PEFT, providing insights that may inspire future advancements in the field.

\vspace{-0.15in}

\ifCLASSOPTIONcaptionsoff
  \newpage
\fi

\bibliographystyle{ieeetr}

\begin{thebibliography}{100}

\bibitem{xin2024v}
Y.~Xin, S.~Luo, {\em et~al.}, ``V-petl bench: A unified visual parameter-efficient transfer learning benchmark,'' {\em Advances in Neural Information Processing Systems (NeurIPS)}, 2024.

\bibitem{deng2009imagenet}
J.~Deng, W.~Dong, {\em et~al.}, ``Imagenet: A large-scale hierarchical image database,'' {\em Proceedings of the IEEE Conference on Computer Vision and Pattern Recognition (CVPR)}, 2009.

\bibitem{ridnik2021imagenet}
T.~Ridnik, E.~Ben-Baruch, {\em et~al.}, ``Imagenet-21k pretraining for the masses,'' {\em Advances in Neural Information Processing Systems (NeurIPS)}, 2021.

\bibitem{kay2017kinetics}
W.~Kay, J.~Carreira, {\em et~al.}, ``The kinetics human action video dataset,'' {\em {{Arxiv}}}, 2017.

\bibitem{he2016deep}
K.~He, X.~Zhang, {\em et~al.}, ``Deep residual learning for image recognition,'' {\em Proceedings of the IEEE Conference on Computer Vision and Pattern Recognition (CVPR)}, 2016.

\bibitem{liu2021swin}
Z.~Liu, Y.~Lin, {\em et~al.}, ``Swin transformer: Hierarchical vision transformer using shifted windows,'' {\em Proceedings of the IEEE International Conference on Computer Vision (ICCV)}, 2021.

\bibitem{Wang2021PyramidVT}
H.~Wu, B.~Xiao, {\em et~al.}, ``Pyramid vision transformer: A versatile backbone for dense prediction without convolutions,'' {\em Proceedings of the IEEE International Conference on Computer Vision (ICCV)}, 2021.

\bibitem{kirillov2023segment}
A.~Kirillov, E.~Mintun, {\em et~al.}, ``Segment anything,'' {\em Proceedings of the IEEE International Conference on Computer Vision (ICCV)}, 2023.

\bibitem{touvron2021training}
H.~Touvron, M.~Cord, {\em et~al.}, ``Training data-efficient image transformers \& distillation through attention,'' {\em Proceedings of the International Conference on Machine Learning (ICML)}, 2021.

\bibitem{he2022masked}
K.~He, X.~Chen, {\em et~al.}, ``Masked autoencoders are scalable vision learners,'' {\em Proceedings of the IEEE Conference on Computer Vision and Pattern Recognition (CVPR)}, 2022.

\bibitem{chen2020simple}
T.~Chen, S.~Kornblith, {\em et~al.}, ``A simple framework for contrastive learning of visual representations,'' {\em Proceedings of the International Conference on Machine Learning (ICML)}, 2020.

\bibitem{he2020momentum}
K.~He, H.~Fan, {\em et~al.}, ``Momentum contrast for unsupervised visual representation learning,'' {\em Proceedings of the IEEE Conference on Computer Vision and Pattern Recognition (CVPR)}, 2020.

\bibitem{dosovitskiy2020vit}
A.~Dosovitskiy, L.~Beyer, {\em et~al.}, ``An image is worth 16x16 words: Transformers for image recognition at scale,'' {\em Proceedings of the International Conference on Learning Representations (ICLR)}, 2021.

\bibitem{hou2024conv2former}
Q.~Hou, C.-Z. Lu, {\em et~al.}, ``Conv2former: A simple transformer-style convnet for visual recognition,'' {\em IEEE Transactions on Pattern Analysis and Machine Intelligence (TPAMI)}, 2024.

\bibitem{li2022contextual}
Y.~Li, T.~Yao, {\em et~al.}, ``Contextual transformer networks for visual recognition,'' {\em IEEE Transactions on Pattern Analysis and Machine Intelligence (TPAMI)}, 2022.

\bibitem{yang2022recurring}
J.~Yang, X.~Dong, {\em et~al.}, ``Recurring the transformer for video action recognition,'' {\em Proceedings of the IEEE Conference on Computer Vision and Pattern Recognition (CVPR)}, 2022.

\bibitem{xing2023svformer}
Z.~Xing, Q.~Dai, {\em et~al.}, ``Svformer: Semi-supervised video transformer for action recognition,'' {\em Proceedings of the IEEE Conference on Computer Vision and Pattern Recognition (CVPR)}, 2023.

\bibitem{strudel2021segmenter}
R.~Strudel, R.~Garcia, {\em et~al.}, ``Segmenter: Transformer for semantic segmentation,'' {\em Proceedings of the IEEE International Conference on Computer Vision (ICCV)}, 2021.

\bibitem{zhang2022segvit}
B.~Zhang, Z.~Tian, {\em et~al.}, ``Segvit: Semantic segmentation with plain vision transformers,'' {\em Advances in Neural Information Processing Systems (NeurIPS)}, 2022.

\bibitem{chen2023pixart}
J.~Chen {\em et~al.}, ``Pixart-$alpha$: Fast training of diffusion transformer for photorealistic text-to-image synthesis,'' {\em Proceedings of the International Conference on Learning Representations (ICLR)}, 2023.

\bibitem{liu2024lumina}
D.~Liu, S.~Zhao, {\em et~al.}, ``Lumina-mgpt: Illuminate flexible photorealistic text-to-image generation with multimodal generative pretraining,'' {\em {{Arxiv}}}, 2024.

\bibitem{qin2025lumina}
Q.~Qin, L.~Zhuo, {\em et~al.}, ``Lumina-image 2.0: A unified and efficient image generative framework,'' {\em {{Arxiv}}}, 2025.

\bibitem{houlsby2019parameter}
N.~Houlsby, A.~Giurgiu, {\em et~al.}, ``Parameter-efficient transfer learning for nlp,'' {\em Proceedings of the International Conference on Machine Learning (ICML)}, 2019.

\bibitem{kornblith2019better}
S.~Kornblith, J.~Shlens, {\em et~al.}, ``Do better imagenet models transfer better?,'' {\em Proceedings of the IEEE Conference on Computer Vision and Pattern Recognition (CVPR)}, 2019.

\bibitem{yu2022towards}
B.~X. Yu, J.~Chang, {\em et~al.}, ``Towards a unified view on visual parameter-efficient transfer learning,'' {\em {{Arxiv}}}, 2022.

\bibitem{ding2022delta}
N.~Ding, Y.~Qin, {\em et~al.}, ``Delta tuning: A comprehensive study of parameter efficient methods for pre-trained language models,'' {\em Nature Machine Intelligence}, 2022.

\bibitem{xu2023parameter}
L.~Xu, H.~Xie, {\em et~al.}, ``Parameter-efficient fine-tuning methods for pretrained language models: A critical review and assessment,'' {\em {{Arxiv}}}, 2023.

\bibitem{han2024parameter}
Z.~Han, C.~Gao, {\em et~al.}, ``Parameter-efficient fine-tuning for large models: A comprehensive survey,'' {\em {{Arxiv}}}, 2024.

\bibitem{wang2024parameter}
L.~Wang, S.~Chen, {\em et~al.}, ``Parameter-efficient fine-tuning in large models: A survey of methodologies,'' {\em {{Arxiv}}}, 2024.

\bibitem{yuan2021tokens}
L.~Yuan, Y.~Chen, {\em et~al.}, ``Tokens-to-token vit: Training vision transformers from scratch on imagenet,'' {\em Proceedings of the IEEE International Conference on Computer Vision (ICCV)}, 2021.

\bibitem{khan2022transformers}
S.~Khan, M.~Naseer, {\em et~al.}, ``Transformers in vision: A survey,'' {\em ACM Computing Surveys (CSUR)}, 2022.

\bibitem{NEURIPS2020_4c5bcfec}
J.~Ho, A.~Jain, {\em et~al.}, ``Denoising diffusion probabilistic models,'' {\em Advances in Neural Information Processing Systems (NeurIPS)}, 2020.

\bibitem{rombach2022high}
R.~Rombach, A.~Blattmann, {\em et~al.}, ``High-resolution image synthesis with latent diffusion models,'' {\em Proceedings of the IEEE Conference on Computer Vision and Pattern Recognition (CVPR)}, 2022.

\bibitem{podell2023sdxl}
D.~Podell, Z.~English, {\em et~al.}, ``Sdxl: Improving latent diffusion models for high-resolution image synthesis,'' {\em Proceedings of the International Conference on Learning Representations (ICLR)}, 2023.

\bibitem{kingma2013auto}
D.~P. Kingma, M.~Welling, {\em et~al.}, ``Auto-encoding variational bayes,'' {\em {{Arxiv}}}, 2013.

\bibitem{ronneberger2015u}
O.~Ronneberger, P.~Fischer, {\em et~al.}, ``U-net: Convolutional networks for biomedical image segmentation,'' {\em Medical Image Computing and Computer Assisted Intervention (MICCAI)}, 2015.

\bibitem{croitoru2023diffusion}
F.-A. Croitoru, V.~Hondru, {\em et~al.}, ``Diffusion models in vision: A survey,'' {\em IEEE Transactions on Pattern Analysis and Machine Intelligence (TPAMI)}, 2023.

\bibitem{chen2022adaptformer}
S.~Chen, C.~Ge, {\em et~al.}, ``Adaptformer: Adapting vision transformers for scalable visual recognition,'' {\em Advances in Neural Information Processing Systems (NeurIPS)}, 2022.

\bibitem{guo2024i2v}
X.~Guo, M.~Zheng, {\em et~al.}, ``I2v-adapter: A general image-to-video adapter for diffusion models,'' {\em ACM SIGGRAPH 2024 Conference Papers}, 2024.

\bibitem{jie2022convolutional}
S.~Jie and Z.-H. Deng, ``Convolutional bypasses are better vision transformer adapters,'' {\em {{Arxiv}}}, 2022.

\bibitem{sharma2023lossless}
M.~Sharma, C.~Fantacci, {\em et~al.}, ``Lossless adaptation of pretrained vision models for robotic manipulation,'' {\em Proceedings of the International Conference on Learning Representations (ICLR)}, 2023.

\bibitem{mou2024t2i}
C.~Mou {\em et~al.}, ``T2i-adapter: Learning adapters to dig out more controllable ability for text-to-image diffusion models,'' {\em Proceedings of the AAAI Conference on Artificial Intelligence (AAAI)}, 2024.

\bibitem{pan2022st}
J.~Pan, Z.~Lin, {\em et~al.}, ``St-adapter: Parameter-efficient image-to-video transfer learning,'' {\em Advances in Neural Information Processing Systems (NeurIPS)}, 2022.

\bibitem{chen2024artadapter}
D.-Y. Chen, H.~Tennent, {\em et~al.}, ``Artadapter: Text-to-image style transfer using multi-level style encoder and explicit adaptation,'' {\em Proceedings of the IEEE Conference on Computer Vision and Pattern Recognition (CVPR)}, 2024.

\bibitem{yang2023aim}
T.~Yang, Y.~Zhu, {\em et~al.}, ``Aim: Adapting image models for efficient video action recognition,'' {\em Proceedings of the International Conference on Learning Representations (ICLR)}, 2023.

\bibitem{ye2023ip-adapter}
H.~Ye, J.~Zhang, {\em et~al.}, ``Ip-adapter: Text compatible image prompt adapter for text-to-image diffusion models,'' {\em Proceedings of the AAAI Conference on Artificial Intelligence (AAAI)}, 2024.

\bibitem{xing2024simda}
Z.~Xing, Q.~Dai, {\em et~al.}, ``Simda: Simple diffusion adapter for efficient video generation,'' {\em Proceedings of the IEEE Conference on Computer Vision and Pattern Recognition (CVPR)}, 2024.

\bibitem{yin20231}
D.~Yin, Y.~Yang, {\em et~al.}, ``1\% vs 100\%: Parameter-efficient low rank adapter for dense predictions,'' {\em Proceedings of the IEEE Conference on Computer Vision and Pattern Recognition (CVPR)}, 2023.

\bibitem{zhao2023sct}
H.~H. Zhao, P.~Wang, {\em et~al.}, ``Sct: A simple baseline for parameter-efficient fine-tuning via salient channels,'' {\em International Journal of Computer Vision (IJCV)}, 2023.

\bibitem{dong2024efficient}
W.~Dong, D.~Yan, {\em et~al.}, ``Efficient adaptation of large vision transformer via adapter re-composing,'' {\em Advances in Neural Information Processing Systems (NeurIPS)}, 2024.

\bibitem{deng2023selective}
X.~Deng, Q.~Fan, {\em et~al.}, ``Selective feature adapter for dense vision transformers,'' {\em {{Arxiv}}}, 2023.

\bibitem{zhang2023mosa}
Q.~Zhang, B.~Zou, {\em et~al.}, ``Mosa: Mixture of sparse adapters for visual efficient tuning,'' {\em {{Arxiv}}}, 2023.

\bibitem{nowak2024towards}
A.~I. Nowak, O.-B. Mercea, {\em et~al.}, ``Towards optimal adapter placement for efficient transfer learning,'' {\em {{Arxiv}}}, 2024.

\bibitem{marouf2024mini}
I.~E. Marouf, E.~Tartaglione, {\em et~al.}, ``Mini but mighty: Finetuning vits with mini adapters,'' {\em Proceedings of the IEEE Winter Conference on Applications of Computer Vision (WACV)}, 2024.

\bibitem{xu2024memory}
X.~Xu, C.~Xia, {\em et~al.}, ``Memory-based adapters for online 3d scene perception,'' {\em Proceedings of the IEEE Conference on Computer Vision and Pattern Recognition (CVPR)}, 2024.

\bibitem{lei2023conditional}
T.~Lei, J.~Bai, {\em et~al.}, ``Conditional adapters: Parameter-efficient transfer learning with fast inference,'' {\em Advances in Neural Information Processing Systems (NeurIPS)}, 2023.

\bibitem{jia2022visual}
M.~Jia, L.~Tang, {\em et~al.}, ``Visual prompt tuning,'' {\em Proceedings of the European Conference on Computer Vision (ECCV)}, 2022.

\bibitem{tsai2023convolutional}
Y.-Y. Tsai, C.~Mao, {\em et~al.}, ``Convolutional visual prompt for robust visual perception,'' {\em Advances in Neural Information Processing Systems (NeurIPS)}, 2023.

\bibitem{dong2022lpt}
B.~Dong, P.~Zhou, {\em et~al.}, ``Lpt: Long-tailed prompt tuning for image classification,'' {\em Proceedings of the International Conference on Learning Representations (ICLR)}, 2023.

\bibitem{han20232}
C.~Han, Q.~Wang, {\em et~al.}, ``E2vpt: An effective and efficient approach for visual prompt tuning,'' {\em Proceedings of the IEEE International Conference on Computer Vision (ICCV)}, 2023.

\bibitem{zhu2023visual}
J.~Zhu, S.~Lai, {\em et~al.}, ``Visual prompt multi-modal tracking,'' {\em Proceedings of the IEEE Conference on Computer Vision and Pattern Recognition (CVPR)}, 2023.

\bibitem{zha2023instance}
Y.~Zha, J.~Wang, {\em et~al.}, ``Instance-aware dynamic prompt tuning for pre-trained point cloud models,'' {\em Proceedings of the IEEE International Conference on Computer Vision (ICCV)}, 2023.

\bibitem{wang2023lion}
H.~Wang {\em et~al.}, ``Lion: Implicit vision prompt tuning,'' {\em Proceedings of the AAAI Conference on Artificial Intelligence (AAAI)}, 2024.

\bibitem{gao2022visual}
Y.~Gao, X.~Shi, {\em et~al.}, ``Visual prompt tuning for test-time domain adaptation,'' {\em {{Arxiv}}}, 2022.

\bibitem{pei2024sa2vp}
W.~Pei, T.~Xia, {\em et~al.}, ``Sa$^2$vp: Spatially aligned-and-adapted visual prompt,'' {\em Proceedings of the AAAI Conference on Artificial Intelligence (AAAI)}, 2024.

\bibitem{das2023learning}
R.~Das, Y.~Dukler, {\em et~al.}, ``Learning expressive prompting with residuals for vision transformers,'' {\em Proceedings of the IEEE Conference on Computer Vision and Pattern Recognition (CVPR)}, 2023.

\bibitem{park2024fair}
S.~Park and H.~Byun, ``Fair-vpt: Fair visual prompt tuning for image classification,'' {\em Proceedings of the IEEE Conference on Computer Vision and Pattern Recognition (CVPR)}, 2024.

\bibitem{zhu2024semantic}
H.~Zhu, F.~Zhang, {\em et~al.}, ``Semantic hierarchical prompt tuning for parameter-efficient fine-tuning,'' {\em {{Arxiv}}}, 2024.

\bibitem{sohn2023visual}
K.~Sohn, H.~Chang, {\em et~al.}, ``Visual prompt tuning for generative transfer learning,'' {\em Proceedings of the IEEE Conference on Computer Vision and Pattern Recognition (CVPR)}, 2023.

\bibitem{nie2023pro}
X.~Nie, B.~Ni, {\em et~al.}, ``Pro-tuning: Unified prompt tuning for vision tasks,'' {\em IEEE Transactions on Circuits and Systems for Video Technology (TCSVT)}, 2023.

\bibitem{bahng2022exploring}
H.~Bahng, A.~Jahanian, {\em et~al.}, ``Exploring visual prompts for adapting large-scale models,'' {\em {{{Arxiv}}}}, 2022.

\bibitem{liu2023explicit}
W.~Liu, X.~Shen, {\em et~al.}, ``Explicit visual prompting for low-level structure segmentations,'' {\em Proceedings of the IEEE Conference on Computer Vision and Pattern Recognition (CVPR)}, 2023.

\bibitem{wang2022p2p}
Z.~Wang, X.~Yu, {\em et~al.}, ``P2p: Tuning pre-trained image models for point cloud analysis with point-to-pixel prompting,'' {\em Advances in Neural Information Processing Systems (NeurIPS)}, 2022.

\bibitem{wu2022unleashing}
J.~Wu, X.~Li, {\em et~al.}, ``Unleashing the power of visual prompting at the pixel level,'' {\em Transactions on Machine Learning Research (TMLR)}, 2022.

\bibitem{chen2023understanding}
A.~Chen, Y.~Yao, {\em et~al.}, ``Understanding and improving visual prompting: A label-mapping perspective,'' {\em Proceedings of the IEEE Conference on Computer Vision and Pattern Recognition (CVPR)}, 2023.

\bibitem{hu2022prosfda}
S.~Hu, Z.~Liao, {\em et~al.}, ``Prosfda: Prompt learning based source-free domain adaptation for medical image segmentation,'' {\em {{Arxiv}}}, 2022.

\bibitem{huang2023diversity}
Q.~Huang, X.~Dong, {\em et~al.}, ``Diversity-aware meta visual prompting,'' {\em Proceedings of the IEEE Conference on Computer Vision and Pattern Recognition (CVPR)}, 2023.

\bibitem{xu2023exploring}
C.~Xu, S.~Yang, {\em et~al.}, ``Exploring efficient few-shot adaptation for vision transformers,'' {\em Transactions on Machine Learning Research (TMLR)}, 2023.

\bibitem{gao2023unified}
Q.~Gao, C.~Zhao, {\em et~al.}, ``A unified continual learning framework with general parameter-efficient tuning,'' {\em Proceedings of the IEEE International Conference on Computer Vision (ICCV)}, 2023.

\bibitem{tu2023visual}
C.-H. Tu, Z.~Mai, {\em et~al.}, ``Visual query tuning: Towards effective usage of intermediate representations for parameter and memory efficient transfer learning,'' {\em Proceedings of the IEEE Conference on Computer Vision and Pattern Recognition (CVPR)}, 2023.

\bibitem{li2021prefix}
X.~L. Li and P.~Liang, ``Prefix-tuning: Optimizing continuous prompts for generation,'' {\em Proceedings of the Annual Meeting of the Association for Computational Linguistics (ACL)}, 2021.

\bibitem{bandara2024attention}
W.~G.~C. Bandara and V.~M. Patel, ``Attention prompt tuning: Parameter-efficient adaptation of pre-trained models for action recognition,'' {\em International Conference on Automatic Face and Gesture Recognition (FG)}, 2024.

\bibitem{sun2023fedperfix}
G.~Sun, M.~Mendieta, {\em et~al.}, ``Fedperfix: Towards partial model personalization of vision transformers in federated learning,'' {\em Proceedings of the IEEE International Conference on Computer Vision (ICCV)}, 2023.

\bibitem{zhang2023adding}
L.~Zhang, A.~Rao, {\em et~al.}, ``Adding conditional control to text-to-image diffusion models,'' {\em Proceedings of the IEEE Conference on Computer Vision and Pattern Recognition (CVPR)}, 2023.

\bibitem{xu2023side}
M.~Xu, Z.~Zhang, {\em et~al.}, ``Side adapter network for open-vocabulary semantic segmentation,'' {\em Proceedings of the IEEE Conference on Computer Vision and Pattern Recognition (CVPR)}, 2023.

\bibitem{lin2023hierarchical}
W.~Lin, Z.~Wu, {\em et~al.}, ``Hierarchical side-tuning for vision transformers,'' {\em {{Arxiv}}}, 2023.

\bibitem{zhao2024uni}
S.~Zhao, D.~Chen, {\em et~al.}, ``Uni-controlnet: All-in-one control to text-to-image diffusion models,'' {\em Advances in Neural Information Processing Systems (NeurIPS)}, 2024.

\bibitem{zhang2020side}
J.~O. Zhang, A.~Sax, {\em et~al.}, ``Side-tuning: a baseline for network adaptation via additive side networks,'' {\em Proceedings of the European Conference on Computer Vision (ECCV)}, 2020.

\bibitem{chen2022vision}
Z.~Chen, Y.~Duan, {\em et~al.}, ``Vision transformer adapter for dense predictions,'' {\em Proceedings of the International Conference on Learning Representations (ICLR)}, 2023.

\bibitem{chai2023ladder}
S.~Chai, R.~K. Jain, {\em et~al.}, ``Ladder fine-tuning approach for sam integrating complementary network,'' {\em {{Arxiv}}}, 2023.

\bibitem{yin2023parameter}
D.~Yin, X.~Han, {\em et~al.}, ``Parameter-efficient is not sufficient: Exploring parameter, memory, and time efficient adapter tuning for dense predictions,'' {\em {{Arxiv}}}, 2023.

\bibitem{mercea2024time}
O.-B. Mercea, A.~Gritsenko, C.~Schmid, and A.~Arnab, ``Time-memory-and parameter-efficient visual adaptation,'' {\em Proceedings of the IEEE Conference on Computer Vision and Pattern Recognition (CVPR)}, 2024.

\bibitem{sung2022lst}
Y.-L. Sung, J.~Cho, {\em et~al.}, ``Lst: Ladder side-tuning for parameter and memory efficient transfer learning,'' {\em Advances in Neural Information Processing Systems (NeurIPS)}, 2022.

\bibitem{fu2023dtl}
M.~Fu, K.~Zhu, {\em et~al.}, ``Dtl: Disentangled transfer learning for visual recognition,'' {\em Proceedings of the AAAI Conference on Artificial Intelligence (AAAI)}, 2024.

\bibitem{fu2022adapterbias}
C.-L. Fu, Z.-C. Chen, {\em et~al.}, ``Adapterbias: Parameter-efficient token-dependent representation shift for adapters in nlp tasks,'' {\em Proceedings of the Annual Conference of the North American Chapter of the Association for Computational Linguistics (NAACL)}, 2022.

\bibitem{xie2023difffit}
E.~Xie, L.~Yao, {\em et~al.}, ``Difffit: Unlocking transferability of large diffusion models via simple parameter-efficient fine-tuning,'' {\em Proceedings of the IEEE International Conference on Computer Vision (ICCV)}, 2023.

\bibitem{bu2022differentially}
Z.~Bu, Y.-X. Wang, {\em et~al.}, ``Differentially private bias-term only fine-tuning of foundation models,'' {\em Advances in Neural Information Processing Systems (NeurIPS)}, 2022.

\bibitem{basu2023strong}
S.~Basu, S.~Hu, {\em et~al.}, ``Strong baselines for parameter efficient few-shot fine-tuning,'' {\em Proceedings of the AAAI Conference on Artificial Intelligence (AAAI)}, 2024.

\bibitem{zhang2024gradient}
Z.~Zhang, Q.~Zhang, {\em et~al.}, ``Gradient-based parameter selection for efficient fine-tuning,'' {\em Proceedings of the IEEE Conference on Computer Vision and Pattern Recognition (CVPR)}, 2024.

\bibitem{zaken2021bitfit}
E.~B. Zaken, S.~Ravfogel, {\em et~al.}, ``Bitfit: Simple parameter-efficient fine-tuning for transformer-based masked language-models,'' {\em Proceedings of the Annual Meeting of the Association for Computational Linguistics (ACL)}, 2022.

\bibitem{tamirisa2024fedselect}
R.~Tamirisa, C.~Xie, {\em et~al.}, ``Fedselect: Personalized federated learning with customized selection of parameters for fine-tuning,'' {\em Proceedings of the IEEE Conference on Computer Vision and Pattern Recognition (CVPR)}, 2024.

\bibitem{hu2021lora}
E.~J. Hu, Y.~Shen, {\em et~al.}, ``Lora: Low-rank adaptation of large language models,'' {\em Proceedings of the International Conference on Learning Representations (ICLR)}, 2021.

\bibitem{edalati2022krona}
A.~Edalati, M.~Tahaei, {\em et~al.}, ``Krona: Parameter efficient tuning with kronecker adapter,'' {\em {{Arxiv}}}, 2022.

\bibitem{jie2023fact}
S.~Jie and Z.-H. Deng, ``Fact: Factor-tuning for lightweight adaptation on vision transformer,'' {\em Proceedings of the AAAI Conference on Artificial Intelligence (AAAI)}, 2023.

\bibitem{chen2023aggregate}
D.~Chen, ``Aggregate, decompose, and fine-tune: A simple yet effective factor-tuning method for vision transformer,'' {\em {{Arxiv}}}, 2023.

\bibitem{lian2022scaling}
D.~Lian, D.~Zhou, {\em et~al.}, ``Scaling \& shifting your features: A new baseline for efficient model tuning,'' {\em Advances in Neural Information Processing Systems (NeurIPS)}, 2022.

\bibitem{jiang2022dna}
Z.~Jiang, T.~Chen, {\em et~al.}, ``Dna: Improving few-shot transfer learning with low-rank decomposition and alignment,'' {\em Proceedings of the European Conference on Computer Vision (ECCV)}, 2022.

\bibitem{du2024alore}
S.~Du, G.~Zhang, {\em et~al.}, ``Alore: Efficient visual adaptation via aggregating low rank experts,'' {\em {{Arxiv}}}, 2024.

\bibitem{he2023parameter}
X.~He, C.~Li, {\em et~al.}, ``Parameter-efficient model adaptation for vision transformers,'' {\em Proceedings of the AAAI Conference on Artificial Intelligence (AAAI)}, 2023.

\bibitem{fang2025dropout}
Z.~Fang, Y.~Wang, {\em et~al.}, ``Dropout mixture low-rank adaptation for visual parameters-efficient fine-tuning,'' {\em Proceedings of the European Conference on Computer Vision (ECCV)}, 2025.

\bibitem{gandikota2024concept}
R.~Gandikota, J.~Materzy{\'n}ska, {\em et~al.}, ``Concept sliders: Lora adaptors for precise control in diffusion models,'' {\em Proceedings of the European Conference on Computer Vision (ECCV)}, 2024.

\bibitem{luo2023towards}
G.~Luo, M.~Huang, {\em et~al.}, ``Towards efficient visual adaption via structural re-parameterization,'' {\em {{Arxiv}}}, 2023.

\bibitem{liu2022neural}
Y.~Zhang, K.~Zhou, {\em et~al.}, ``Neural prompt search,'' {\em IEEE Transactions on Pattern Analysis and Machine Intelligence (TPAMI)}, 2024.

\bibitem{jiang2023rethinking}
Z.~Jiang, C.~Mao, {\em et~al.}, ``Rethinking efficient tuning methods from a unified perspective,'' {\em {{Arxiv}}}, 2023.

\bibitem{zhou2024dynamic}
X.~Zhou, D.~Liang, {\em et~al.}, ``Dynamic adapter meets prompt tuning: Parameter-efficient transfer learning for point cloud analysis,'' {\em Proceedings of the IEEE Conference on Computer Vision and Pattern Recognition (CVPR)}, 2024.

\bibitem{mai2024lessons}
Z.~Mai, P.~Zhang, {\em et~al.}, ``Lessons learned from a unifying empirical study of parameter-efficient transfer learning (petl) in visual recognition,'' {\em Proceedings of the IEEE Conference on Computer Vision and Pattern Recognition (CVPR)}, 2025.

\bibitem{zhao2024dynamic}
W.~Zhao, J.~Tang, {\em et~al.}, ``Dynamic tuning towards parameter and inference efficiency for vit adaptation,'' {\em Advances in Neural Information Processing Systems (NeurIPS)}, 2024.

\bibitem{liu2024sparse}
T.~Liu, X.~Liu, {\em et~al.}, ``Sparse-tuning: Adapting vision transformers with efficient fine-tuning and inference,'' {\em {{Arxiv}}}, 2024.

\bibitem{ly2024enhancing}
S.~T. Ly and H.~V. Nguyen, ``Enhancing parameter-efficient fine-tuning of vision transformers through frequency-based adaptation,'' {\em {{Arxiv}}}, 2024.

\bibitem{dongefficient}
W.~Dong, Y.~Sun, {\em et~al.}, ``Efficient adaptation of pre-trained vision transformer via householder transformation,'' {\em Advances in Neural Information Processing Systems (NeurIPS)}, 2024.

\bibitem{ruan2024gist}
J.~Ruan, J.~Gao, {\em et~al.}, ``Gist: Improving parameter efficient fine-tuning via knowledge interaction,'' {\em Proceedings of the ACM Conference on Multimedia (MM)}, 2024.

\bibitem{zhang2025dyn}
Y.~Zhang, H.~Chen, {\em et~al.}, ``Dyn-adapter: Towards disentangled representation for efficient visual recognition,'' {\em Proceedings of the European Conference on Computer Vision (ECCV)}, 2025.

\bibitem{han2025straightforward}
R.~Han and J.~Tang, ``Straightforward layer-wise pruning for more efficient visual adaptation,'' {\em Proceedings of the European Conference on Computer Vision (ECCV)}, 2025.

\bibitem{liu2022polyhistor}
Y.-C. Liu, C.-Y. Ma, {\em et~al.}, ``Polyhistor: Parameter-efficient multi-task adaptation for dense vision tasks,'' {\em Advances in Neural Information Processing Systems (NeurIPS)}, 2022.

\bibitem{xin2023vmt}
Y.~Xin, J.~Du, {\em et~al.}, ``Vmt-adapter: Parameter-efficient transfer learning for multi-task dense,'' {\em Proceedings of the AAAI Conference on Artificial Intelligence (AAAI)}, 2024.

\bibitem{zhongtransforming}
H.~Zhong, J.~Chen, {\em et~al.}, ``Transforming vision transformer: Towards efficient multi-task asynchronous learner,'' {\em Advances in Neural Information Processing Systems (NeurIPS)}, 2024.

\bibitem{yang2024multi}
Y.~Yang, P.-T. Jiang, {\em et~al.}, ``Multi-task dense prediction via mixture of low-rank experts,'' {\em Proceedings of the IEEE Conference on Computer Vision and Pattern Recognition (CVPR)}, 2024.

\bibitem{zhu2024vl}
M.~Zhu, G.~Liu, {\em et~al.}, ``Vl-mpft: Multitask parameter-efficient fine-tuning for visual-language pre-trained models via task-adaptive masking,'' {\em Chinese Conference on Pattern Recognition and Computer Vision (PRCV)}, 2024.

\bibitem{xin2024mmap}
Y.~Xin, J.~Du, {\em et~al.}, ``Mmap: Multi-modal alignment prompt for cross-domain multi-task learning,'' {\em Proceedings of the AAAI Conference on Artificial Intelligence (AAAI)}, 2024.

\bibitem{mahabadi2021parameter}
R.~K. Mahabadi, S.~Ruder, {\em et~al.}, ``Parameter-efficient multi-task fine-tuning for transformers via shared hypernetworks,'' {\em Proceedings of the Annual Meeting of the Association for Computational Linguistics (ACL)}, 2021.

\bibitem{baek2025tadformer}
S.~Baek, S.~Lee, {\em et~al.}, ``Tadformer: Task-adaptive dynamic transformer for efficient multi-task learning,'' {\em {{Arxiv}}}, 2025.

\bibitem{agiza2024mtlora}
A.~Agiza {\em et~al.}, ``Mtlora: Low-rank adaptation approach for efficient multi-task learning,'' {\em Proceedings of the IEEE Conference on Computer Vision and Pattern Recognition (CVPR)}, 2024.

\bibitem{Caron2021EmergingPI}
M.~Caron, H.~Touvron, {\em et~al.}, ``Emerging properties in self-supervised vision transformers,'' {\em Proceedings of the IEEE International Conference on Computer Vision (ICCV)}, 2021.

\bibitem{Radford2021LearningTV}
A.~Radford, J.~W. Kim, {\em et~al.}, ``Learning transferable visual models from natural language supervision,'' {\em Proceedings of the International Conference on Machine Learning (ICML)}, 2021.

\bibitem{Jia2021ScalingUV}
C.~Jia, Y.~Yang, {\em et~al.}, ``Scaling up visual and vision-language representation learning with noisy text supervision,'' {\em Proceedings of the International Conference on Machine Learning (ICML)}, 2021.

\bibitem{lin2024vila}
J.~Lin, H.~Yin, {\em et~al.}, ``Vila: On pre-training for visual language models,'' {\em Proceedings of the IEEE Conference on Computer Vision and Pattern Recognition (CVPR)}, 2024.

\bibitem{Xie2021SimMIMAS}
Z.~Xie, Z.~Zhang, {\em et~al.}, ``Simmim: a simple framework for masked image modeling,'' {\em Proceedings of the IEEE Conference on Computer Vision and Pattern Recognition (CVPR)}, 2021.

\bibitem{Fang2022EVAET}
Y.~Fang, W.~Wang, {\em et~al.}, ``Eva: Exploring the limits of masked visual representation learning at scale,'' {\em Proceedings of the IEEE Conference on Computer Vision and Pattern Recognition (CVPR)}, 2022.

\bibitem{huang2024context}
L.~Huang, W.~Wang, {\em et~al.}, ``In-context lora for diffusion transformers,'' {\em {{Arxiv}}}, 2024.

\bibitem{wu2024difflora}
Y.~Wu, Y.~Shi, {\em et~al.}, ``Difflora: Generating personalized low-rank adaptation weights with diffusion,'' {\em {{Arxiv}}}, 2024.

\bibitem{lin2024tracking}
L.~Lin, H.~Fan, {\em et~al.}, ``Tracking meets lora: Faster training, larger model, stronger performance,'' {\em Proceedings of the European Conference on Computer Vision (ECCV)}, 2024.

\bibitem{wei2024online}
X.~Wei, G.~Li, {\em et~al.}, ``Online-lora: Task-free online continual learning via low rank adaptation,'' {\em {{Arxiv}}}, 2024.

\bibitem{jiang2024task}
F.~Jiang, S.~Wang, {\em et~al.}, ``Task-conditional adapter for multi-task dense prediction,'' {\em Proceedings of the ACM Conference on Multimedia (MM)}, 2024.

\bibitem{wah2011caltech}
C.~Wah, S.~Branson, {\em et~al.}, ``The caltech-ucsd birds-200-2011 dataset,'' {\em {California Institute of Technology}}, 2011.

\bibitem{van2015building}
G.~Van~Horn, S.~Branson, {\em et~al.}, ``Building a bird recognition app and large scale dataset with citizen scientists: The fine print in fine-grained dataset collection,'' {\em Proceedings of the IEEE Conference on Computer Vision and Pattern Recognition (CVPR)}, 2015.

\bibitem{nilsback2008automated}
M.-E. Nilsback and A.~Zisserman, ``Automated flower classification over a large number of classes,'' {\em 2008 Sixth Indian conference on computer vision, graphics \& image processing}, 2008.

\bibitem{dataset2011novel}
E.~Dataset, ``Novel datasets for fine-grained image categorization,'' {\em Proceedings of the IEEE Conference on Computer Vision and Pattern Recognition Workshops (CVPR Workshops)}, 2011.

\bibitem{gebru2017fine}
T.~Gebru, J.~Krause, {\em et~al.}, ``Fine-grained car detection for visual census estimation,'' {\em Proceedings of the AAAI Conference on Artificial Intelligence (AAAI)}, 2017.

\bibitem{zhai2019large}
X.~Zhai, J.~Puigcerver, {\em et~al.}, ``A large-scale study of representation learning with the visual task adaptation benchmark,'' {\em {{Arxiv}}}, 2019.

\bibitem{krizhevsky2009learning}
A.~Krizhevsky, G.~Hinton, {\em et~al.}, ``Learning multiple layers of features from tiny images,'' {\em https://www.cs.toronto.edu/kriz/learning-features-2009-TR.pdf}, 2009.

\bibitem{fei2006one}
L.~Fei-Fei, R.~Fergus, {\em et~al.}, ``One-shot learning of object categories,'' {\em IEEE Transactions on Pattern Analysis and Machine Intelligence (TPAMI)}, 2006.

\bibitem{cimpoi2014describing}
M.~Cimpoi, S.~Maji, {\em et~al.}, ``Describing textures in the wild,'' {\em Proceedings of the IEEE Conference on Computer Vision and Pattern Recognition (CVPR)}, 2014.

\bibitem{parkhi2012cats}
O.~M. Parkhi {\em et~al.}, ``Cats and dogs,'' {\em Proceedings of the IEEE Conference on Computer Vision and Pattern Recognition (CVPR)}, 2012.

\bibitem{netzer2011reading}
Y.~Netzer, T.~Wang, {\em et~al.}, ``Reading digits in natural images with unsupervised feature learning,'' {\em Advances in Neural Information Processing Systems Workshops (NeurIPS Workshops)}, 2011.

\bibitem{xiao2010sun}
J.~Xiao, J.~Hays, {\em et~al.}, ``Sun database: Large-scale scene recognition from abbey to zoo,'' {\em Proceedings of the IEEE Conference on Computer Vision and Pattern Recognition (CVPR)}, 2010.

\bibitem{veeling2018rotation}
B.~S. Veeling, J.~Linmans, {\em et~al.}, ``Rotation equivariant cnns for digital pathology,'' {\em Medical Image Computing and Computer Assisted Intervention (MICCAI)}, 2018.

\bibitem{helber2019eurosat}
P.~Helber, B.~Bischke, {\em et~al.}, ``Eurosat: A novel dataset and deep learning benchmark for land use and land cover classification,'' {\em IEEE Journal of Selected Topics in Applied Earth Observations and Remote Sensing}, 2019.

\bibitem{cheng2017remote}
G.~Cheng, J.~Han, and X.~Lu, ``Remote sensing image scene classification: Benchmark and state of the art,'' {\em Proceedings of the IEEE}, 2017.

\bibitem{graham2015kaggle}
B.~Graham, ``Kaggle diabetic retinopathy detection competition report,'' {\em University of Warwick}, 2015.

\bibitem{johnson2017clevr}
J.~Johnson, B.~Hariharan, {\em et~al.}, ``Clevr: A diagnostic dataset for compositional language and elementary visual reasoning,'' {\em Proceedings of the IEEE Conference on Computer Vision and Pattern Recognition (CVPR)}, 2017.

\bibitem{beattie2016deepmind}
C.~Beattie, J.~Z. Leibo, {\em et~al.}, ``Deepmind lab,'' {\em {{Arxiv}}}, 2016.

\bibitem{geiger2013vision}
A.~Geiger, P.~Lenz, {\em et~al.}, ``Vision meets robotics: The kitti dataset,'' {\em The International Journal of Robotics Research}, 2013.

\bibitem{dsprites17}
L.~Matthey, I.~Higgins, {\em et~al.}, ``dsprites: Disentanglement testing sprites dataset.'' https://github.com/deepmind/dsprites-dataset/, 2017.

\bibitem{lecun2004learning}
Y.~LeCun, F.~J. Huang, {\em et~al.}, ``Learning methods for generic object recognition with invariance to pose and lighting,'' {\em Proceedings of the IEEE Conference on Computer Vision and Pattern Recognition (CVPR)}, 2004.

\bibitem{recht2019imagenet}
B.~Recht, R.~Roelofs, {\em et~al.}, ``Do imagenet classifiers generalize to imagenet?,'' {\em Proceedings of the International Conference on Machine Learning (ICML)}, 2019.

\bibitem{wang2019learning}
H.~Wang, S.~Ge, {\em et~al.}, ``Learning robust global representations by penalizing local predictive power,'' {\em Advances in Neural Information Processing Systems (NeurIPS)}, 2019.

\bibitem{hendrycks2021natural}
D.~Hendrycks, K.~Zhao, {\em et~al.}, ``Natural adversarial examples,'' {\em Proceedings of the IEEE Conference on Computer Vision and Pattern Recognition (CVPR)}, 2021.

\bibitem{hendrycks2021many}
D.~Hendrycks {\em et~al.}, ``The many faces of robustness: A critical analysis of out-of-distribution generalization,'' {\em Proceedings of the IEEE International Conference on Computer Vision (ICCV)}, 2021.

\bibitem{carreira2019short}
J.~Carreira, E.~Noland, {\em et~al.}, ``A short note on the kinetics-700 human action dataset,'' {\em {{Arxiv}}}, 2019.

\bibitem{goyal2017something}
R.~Goyal, S.~Ebrahimi~Kahou, {\em et~al.}, ``The" something something" video database for learning and evaluating visual common sense,'' {\em Proceedings of the IEEE International Conference on Computer Vision (ICCV)}, 2017.

\bibitem{kuehne2011hmdb}
H.~Kuehne, H.~Jhuang, {\em et~al.}, ``Hmdb: a large video database for human motion recognition,'' {\em Proceedings of the IEEE International Conference on Computer Vision (ICCV)}, 2011.

\bibitem{soomro2012ucf101}
K.~Soomro and et~al., ``Ucf101: A dataset of 101 human actions classes from videos in the wild,'' {\em {{Arxiv}}}, 2012.

\bibitem{li2018resound}
Y.~Li, Y.~Li, {\em et~al.}, ``Resound: Towards action recognition without representation bias,'' {\em Proceedings of the European Conference on Computer Vision (ECCV)}, 2018.

\bibitem{damen2022rescaling}
D.~Damen, H.~Doughty, {\em et~al.}, ``Rescaling egocentric vision: Collection, pipeline and challenges for epic-kitchens-100,'' {\em International Journal of Computer Vision (IJCV)}, 2022.

\bibitem{lin2014microsoft}
T.-Y. Lin, M.~Maire, {\em et~al.}, ``Microsoft coco: Common objects in context,'' {\em Proceedings of the European Conference on Computer Vision (ECCV)}, 2014.

\bibitem{zhou2019semantic}
B.~Zhou, H.~Zhao, {\em et~al.}, ``Semantic understanding of scenes through the ade20k dataset,'' {\em International Journal of Computer Vision (IJCV)}, 2019.

\bibitem{everingham2015pascal}
M.~Everingham, S.~A. Eslami, {\em et~al.}, ``The pascal visual object classes challenge: A retrospective,'' {\em International Journal of Computer Vision (IJCV)}, 2015.

\bibitem{ma2021abdomenct}
J.~Ma, Y.~Zhang, {\em et~al.}, ``Abdomenct-1k: Is abdominal organ segmentation a solved problem?,'' {\em IEEE Transactions on Pattern Analysis and Machine Intelligence (TPAMI)}, 2021.

\bibitem{antonelli2022medical}
M.~Antonelli, A.~Reinke, {\em et~al.}, ``The medical segmentation decathlon,'' {\em Nature communications}, 2022.

\bibitem{al2020dataset}
W.~Al-Dhabyani, M.~Gomaa, {\em et~al.}, ``Dataset of breast ultrasound images,'' {\em Data in brief}, 2020.

\bibitem{luo2025segrap2023}
X.~Luo, J.~Fu, {\em et~al.}, ``Segrap2023: A benchmark of organs-at-risk and gross tumor volume segmentation for radiotherapy planning of nasopharyngeal carcinoma,'' {\em Medical Image Analysis}, 2025.

\bibitem{zhou2017scene}
B.~Zhou, H.~Zhao, {\em et~al.}, ``Scene parsing through ade20k dataset,'' {\em Proceedings of the IEEE Conference on Computer Vision and Pattern Recognition (CVPR)}, 2017.

\bibitem{peng2024parameter}
Z.~Peng, Z.~Xu, {\em et~al.}, ``Parameter efficient fine-tuning via cross block orchestration for segment anything model,'' {\em Proceedings of the IEEE Conference on Computer Vision and Pattern Recognition (CVPR)}, 2024.

\bibitem{kim2023hydra}
S.~Kim, H.~Yang, {\em et~al.}, ``Hydra: Multi-head low-rank adaptation for parameter efficient fine-tuning,'' {\em {{Arxiv}}}, 2023.

\bibitem{fu2024dtl}
M.~Fu, K.~Zhu, {\em et~al.}, ``Dtl: Disentangled transfer learning for visual recognition,'' {\em Proceedings of the AAAI Conference on Artificial Intelligence (AAAI)}, 2024.

\bibitem{tang2024low}
N.~Tang, M.~Fu, {\em et~al.}, ``Low-rank attention side-tuning for parameter-efficient fine-tuning,'' {\em {{Arxiv}}}, 2024.

\bibitem{wang2023adapting}
Y.~Wang, B.~Shi, {\em et~al.}, ``Adapting shortcut with normalizing flow: An efficient tuning framework for visual recognition,'' {\em Proceedings of the IEEE Conference on Computer Vision and Pattern Recognition (CVPR)}, 2023.

\bibitem{basu2024strong}
S.~Basu, S.~Hu, {\em et~al.}, ``Strong baselines for parameter-efficient few-shot fine-tuning,'' {\em Proceedings of the AAAI Conference on Artificial Intelligence (AAAI)}, 2024.

\bibitem{yin2023adapter}
D.~Yin, L.~Li, {\em et~al.}, ``Adapter is all you need for tuning visual tasks,'' {\em {{Arxiv}}}, 2023.

\bibitem{tong2022videomae}
Z.~Tong, Y.~Song, {\em et~al.}, ``Videomae: Masked autoencoders are data-efficient learners for self-supervised video pre-training,'' {\em Advances in Neural Information Processing Systems (NeurIPS)}, 2022.

\bibitem{liu2022video}
Z.~Liu, J.~Ning, {\em et~al.}, ``Video swin transformer,'' {\em Proceedings of the IEEE Conference on Computer Vision and Pattern Recognition (CVPR)}, 2022.

\bibitem{lilarge}
X.~Li, F.~Tramer, {\em et~al.}, ``Large language models can be strong differentially private learners,'' {\em Proceedings of the International Conference on Learning Representations (ICLR)}, 2012.

\end{thebibliography}

\end{document}